\ifx\synctex\undefined\else
\fi
\ifx\pdfminorversion\undefined\else
\fi
\documentclass[twoside,11pt]{article}

\PassOptionsToPackage{hypertexnames=false,hyperfootnotes=false}{hyperref}
\usepackage{jmlr2e}
\hypersetup{hidelinks}
\usepackage{url}
\usepackage{amsmath}
\usepackage{algorithm}
\usepackage{algpseudocode}
\usepackage{tikz-cd}
\usepackage{booktabs}
\usepackage{multirow}
\usepackage{threeparttable}
\usepackage{adjustbox}
\newtheorem{assumption}[theorem]{Assumption}

\usepackage{lastpage}
\jmlrheading{}{2026}{}{}{}{}{Zhangyong Liang, Ji Zhang, and Huanhuan Gao}

\ShortHeadings{Neural Dynamic Data Valuation}{Liang, Zhang, and Gao}
\firstpageno{1}

\begin{document}

\title{Neural Dynamic Data Valuation via Stochastic State-Adjoint Trajectories}

\author{\name Zhangyong Liang \email liangzhangyong1994@gmail.com \\
       \addr National Center for Applied Mathematics \\ 
       Tianjin University\\ Tianjin 300072, China
       \AND
       \name Ji Zhang \email ji.zhang@unisq.edu.au \\
       \addr School of Mathematics, Physics and Computing\\
       University of Southern Queensland\\ Toowoomba, Queensland 4350, Australia
       \AND
       \name Huanhuan Gao\thanks{Corresponding author.} \email gao\_huanhuan@jlu.edu.cn \\
       \addr School of Mechanical and Aerospace Engineering \\
       Jilin University \\ Changchun 130025, China }

\editor{}

\maketitle

\begin{abstract}
Classical data valuation defines a data point's value through the finite marginal contribution $U(C\cup\{i\})-U(C)$, but estimating this quantity over coalitions requires repeated training and does not describe the contribution made along a stochastic training path.
We ask whether marginal contributions of data points can be estimated from one coupled trajectory while retaining a verifiable relation to coalition-based values.
To this end, we introduce Neural Dynamic Data Valuation (NDDV), which models each data point as a controlled stochastic state and computes a first-order marginal-contribution score via the adjoint equation of the Stochastic Maximum Principle (SMP).
This raw sensitivity is then calibrated by a mass-preserving redistribution that increases one data point's participation while redistributing the same total weight over the remaining data points.
We prove that the resulting backward adjoint recursion is the exact reverse-mode adjoint of the frozen-aggregate Euler system, bound its discrepancy from the mean-field sensitivity, and express each finite coalition marginal as an integral of local sample-weight sensitivities.
These results yield pair-specific error bounds and sufficient conditions for ordering agreement with Shapley, Banzhaf, and leave-one-out values.
Experiments on existing benchmarks evaluate marginal-contribution fidelity, score-release cost, corrupted-sample detection, ablations, and failure regimes.
NDDV is a one-run, trajectory-conditioned estimator, not an unconditional replacement for cooperative-game values.
\end{abstract}


\section{Introduction}
\label{sec:introduction}

Data valuation asks how much each data point contributes to model utility.
The standard formulation begins with a set function $U(C)$: a learning algorithm is trained on a coalition $C$, evaluated on a validation set, and the contribution of data point $i$ is measured by the finite difference $U(C\cup\{i\})-U(C)$.
Leave-one-out~\citep{koh2017understanding}, Shapley~\citep{ghorbani2019,jia2019}, Beta Shapley~\citep{kwon2022beta}, Banzhaf~\citep{wang2022data}, and related semi-values aggregate such marginal contributions over different coalition distributions.
Accurate Shapley estimation often relies on sampled permutations~\citep{mitchell2022sampling}.
Related Shapley explanation work studies interaction indices~\citep{JMLR:v24:22-0202}, SHAP score complexity~\citep{JMLR:v24:21-0389}, and removal-based attribution~\citep{JMLR:v22:20-1316}.
This formulation defines a principled reference value, but computing it accurately usually entails repeated model fitting.

Repeated training, however, is not the only limitation of this formulation.
When the learning algorithm is stochastic, coalition valuation and the contribution to a particular trained model answer different questions.
A coalition utility describes a counterfactual learning algorithm run on another dataset.
A deployed model, by contrast, is produced by one realized sequence of states, mini-batches, and stochastic updates.
Along that path, the effect of a data point can change with training time and with the states of the other samples.
Static retraining omits information about where and how a contribution arose, while a purely local gradient score need not approximate the finite add-one marginal contribution.

The central question is whether each data point's marginal contribution can be estimated from a single stochastic training trajectory, without retraining on data coalitions, and whether checkable conditions can be stated under which this trajectory-based score agrees with finite coalition marginal contributions.

Several lines of prior work address parts of this question.
Influence functions study infinitesimal sample reweighting~\citep{koh2017understanding,feldman2020neural}, and checkpoint methods such as TracIn accumulate gradient interactions along training~\citep{pruthi2020tracin}.
LAVA differentiates a distributional proxy with respect to data mass~\citep{just2023lava}.
GhostSuite and LossVal provide run-specific valuation comparators for training-time data attribution~\citep{wang2024data,wibiral2024lossval}.
Eigen-Value provides an eigenvalue-based route to scalable, domain-robust data valuation~\citep{choi2026eigen}.
These methods establish that sensitivity and run-conditioned attribution can avoid full retraining.
They do not jointly provide a general adjoint formula for a coupled stochastic state system and an explicit error bridge from the implemented trajectory sensitivity to the finite coalition marginal $U(C\cup\{i\})-U(C)$.

We address this problem with \textbf{Neural Dynamic Data Valuation (NDDV)}.
Each data point is represented by a controlled stochastic state, and the states interact through a weighted mean field.
A continuous participation parameter specifies the direction in which a data point enters the learning dynamics.
The Stochastic Maximum Principle (SMP)~\citep{pavliotis2014stochastic} yields an adjoint that converts this direction into a first-order change of the learning criterion.
NDDV uses sensitivity analysis as a computational route to marginal contribution: marginal contribution remains the valuation target, while the adjoint avoids separate retraining for each data point and coalition.

The final score has two stages.
First, the state--adjoint pairing gives a raw coordinate sensitivity for each data point.
Second, we calibrate this sensitivity along a mass-preserving direction that increases the participation of data point $i$ while decreasing the participation of the remaining data points by the same total amount.
This calibration yields the centered NDDV score and gives its subtraction term a variational interpretation rather than treating it as an arbitrary normalization.

The theoretical analysis follows the same question from the population model to the finite marginal contribution.
We first derive an adjoint representation for sample-level marginal directions entering through the initial state, drift, running cost, or terminal objective.
We then show that the recursion implemented by NDDV is the exact reverse-mode adjoint of the sampled, frozen-aggregate Euler system.
A stability result bounds the difference between this system and the population mean-field model.
Finally, a continuous sample-weight path writes the finite coalition marginal contribution as an integral of local sensitivities.
Combining this identity with model, calibration, and coalition-context errors gives a pair-specific bound and a sufficient value-gap condition for agreement with any symmetric semi-value.

The scope of the claim is deliberately narrow.
NDDV is a first-order, trajectory-conditioned estimator of marginal contribution.
It is not a new cooperative-game allocation rule, and it is not asserted to reproduce Shapley or Banzhaf values unconditionally.
Mean-field reweighting and the structured terminal map specify the trajectory on which the sensitivity is evaluated.
The class-conditional quantities used later are auxiliary diagnostics rather than demographic-fairness guarantees.

Our contributions are:
\begin{itemize}
    \item \textbf{A one-run marginal-contribution problem.} We formulate trajectory-conditioned contribution as the sample-participation derivative of a realized stochastic training path and distinguish it from algorithm-level coalition values obtained by independent retraining.
    \item \textbf{Adjoint marginal contribution with mass-preserving calibration.} We derive the sample-level adjoint representation for perturbations entering through the initial state, drift, running cost, or terminal objective.
    The resulting NDDV score is the directional derivative that increases one data point's participation while preserving total data mass.
    \item \textbf{Consistency from the population model to the implemented score.} We separate the population mean-field control problem, the frozen empirical system, and the sampled Euler computation.
    The backward recursion is exact for the discrete frozen-aggregate objective, and stability, discretization, and sampling terms control the gap to the population sensitivity.
    \item \textbf{A verifiable bridge to finite coalition values.} A continuous sample-weight path expresses $U(C\cup\{i\})-U(C)$ as an integral of local sensitivities.
    The resulting pair-specific error budget gives sufficient ordering conditions for Shapley, normalized Banzhaf, and leave-one-out values, together with a finite-sample certificate.
    \item \textbf{Empirical evaluation of fidelity, cost, and downstream use.} Experiments compare NDDV with retraining-based marginal contributions, measure valuation-score release cost, and evaluate corrupted-sample detection and value-directed interventions.
    The results include both regimes in which the ordering condition is informative and regimes in which it is not.
\end{itemize}

Section~\ref{sec:related_works} places NDDV relative to coalition valuation, run-specific attribution, and adjoint learning.
Section~\ref{sec:prelimin} defines the static reference quantities.
Section~\ref{sec:NDDV} develops the one-run trajectory estimator, its adjoint representation, and its connection to coalition values.
Section~\ref{sec:complexity} analyzes computational cost, Section~\ref{sec:experiments} presents the empirical study, and Section~\ref{sec:conclusion} summarizes the conclusions and limitations.

\begin{figure*}[t]
\centering 
\includegraphics[width=\textwidth,height=0.72\textheight,keepaspectratio]{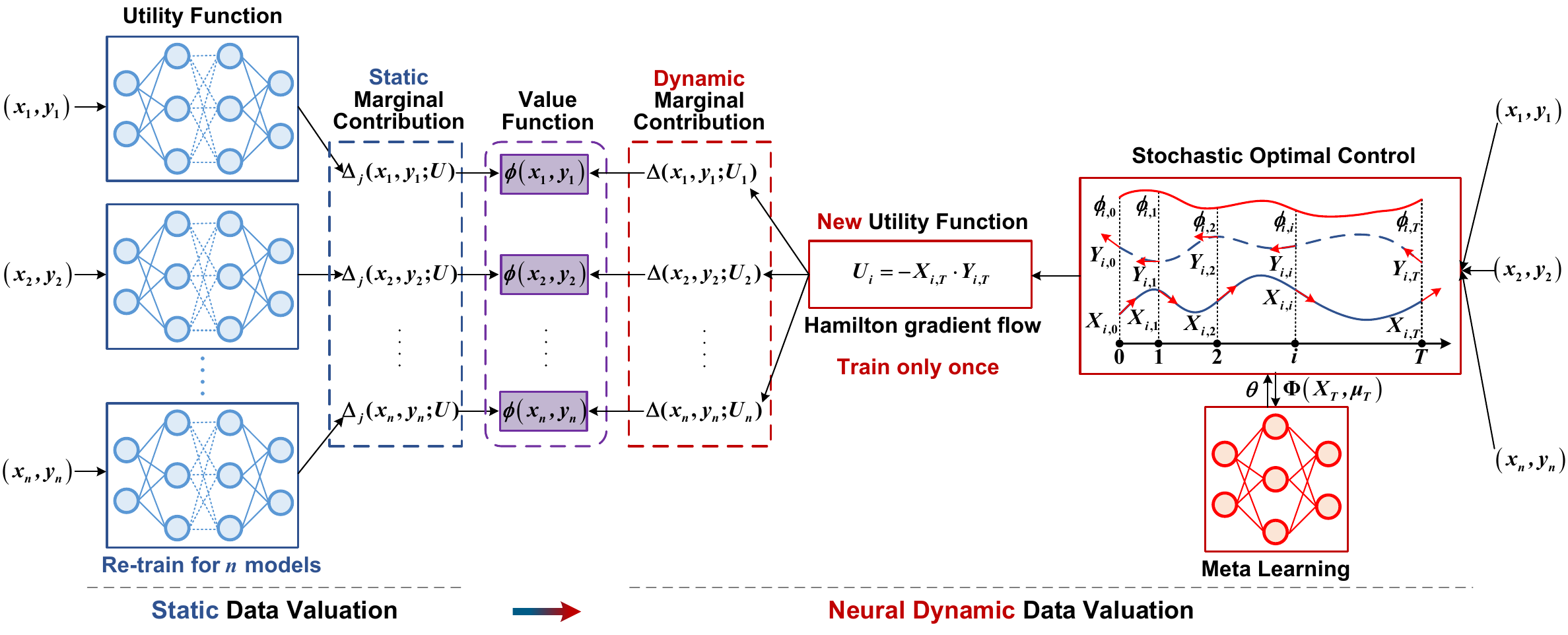}
\vspace{-1em}
\caption{Schematic and empirical overview of NDDV.
The figure summarizes the one-run trajectory valuation pipeline and its comparison with existing data valuation methods.}
\label{fig:nddv}
\end{figure*}

\section{Related Works}
\label{sec:related_works}


\subsection{Coalition Marginal Contribution}

Classical data valuation allocates changes in a set utility to data points.
LOO~\citep{koh2017understanding}, Data Shapley~\citep{ghorbani2019,jia2019}, Beta Shapley~\citep{kwon2022beta}, and Data Banzhaf~\citep{wang2022data} differ in the coalition distribution used to aggregate finite add-one marginals.
They provide important algorithm-level counterfactual targets, but require repeated training or utility evaluation and can be unstable under stochastic retraining.
Scalable variants reduce this cost through nearest-neighbor structure in KNN-Shapley~\citep{jia2019b}, randomized inclusion experiments in AME~\citep{lin2022measuring}, out-of-bag predictions in Data-OOB~\citep{kwon2023dataoob}, sampled utilities in DU-Shapley~\citep{garrido2024shapley}, eigenvalue structure in Eigen-Value~\citep{choi2026eigen}, and transport proxies in LAVA~\citep{just2023lava}.

Shapley-style explanation methods are related but answer a different question.
Permutation sampling improves Shapley estimation~\citep{mitchell2022sampling}, Faith-Shap studies Shapley interaction indices~\citep{JMLR:v24:22-0202}, removal-based attribution evaluates predictive explanations through feature removal~\citep{JMLR:v22:20-1316}, deletion/insertion tests provide related diagnostic criteria~\citep{hama2023deletion}, and SHAP complexity results characterize tractability of explanation scores~\citep{JMLR:v24:21-0389}.

\subsection{Attribution Within a Training Run}

Run-specific attribution avoids explicit coalition retraining by differentiating or tracing a realized optimization process.
Influence functions approximate infinitesimal reweighting effects~\citep{koh2017understanding,feldman2020neural}.
TracIn accumulates checkpoint-gradient interactions~\citep{pruthi2020tracin}, Datamodels learn subset-to-output maps~\citep{ilyas2022datamodels}, TRAK uses projected-gradient features~\citep{park2023trak}, and Gradient$\times$Input rules give local directional sensitivities~\citep{shrikumar2017learning,ancona2019explaining}.
GhostSuite~\citep{wang2024data} and LossVal~\citep{wibiral2024lossval} are recent run-specific data-valuation comparators.

\subsection{Stochastic Control and Adjoint Learning}

Continuous-depth neural models connect residual architectures~\citep{he2016deep} with differential equations and numerical integration~\citep{weinan2017proposal,lu2017beyond}.
Gradient-flow and optimal-transport viewpoints extend this connection to distributions~\citep{JMLR:v20:16-243}, and mean-field control formulations model population-coupled states~\citep{hu2017control,han2018mean}.
The SMP yields the backward adjoint used for sensitivity analysis~\citep{li2017maximum,li2018optimal}, while stochastic gradient-noise models describe related optimization dynamics~\citep{pavliotis2014stochastic,simsekli2019tail}.
Group-aware valuation has been studied through FairShapley-type allocations~\citep{arnaiz2023fairshap,arnaiz2024fairshap} and fairness-aware valuation diagnostics~\citep{pombal2023fairness}.
Decision-level fairness uses different targets, including equalized-odds criteria~\citep{hardt2016equality} and constrained classifiers~\citep{zafar2017fairness}.

\section{Preliminaries}
\label{sec:prelimin}

This section defines the stochastic set utility, finite add-one marginals, and the symmetric semi-values used as static reference quantities. These definitions fix the object that the trajectory-conditioned score is compared against later.

\subsection{Problem Formulation}
Let $\mathcal{D}=\{(x_i,y_i)\}_{i=1}^{N}$ be a supervised training set, where $x_i\in\mathcal X\subset\mathbb R^d$, $y_i\in\mathcal Y$, and $[N]=\{1,\ldots,N\}$.
For a coalition $S\subseteq[N]$, let $\mathcal D_S=\{(x_i,y_i):i\in S\}$.
Let $\xi$ collect the randomness used by the learning algorithm, including initialization, mini-batch order, and injected optimization noise.
Write $\widetilde U(S;\xi)$ for the validation utility obtained by training on $\mathcal D_S$ under a fixed realization $\xi$, and define the algorithm-level higher-is-better utility by
\[
U(S)=\mathbb E_{\xi}[\widetilde U(S;\xi)].
\]
Thus $U:2^{[N]}\to\mathbb R$ is the set value function used by the static cooperative-game reference quantities.
Classical data valuation starts from this set utility: a learning algorithm is trained on $\mathcal D_S$ and evaluated on a fixed validation set.
We take $U(\emptyset)$ to be the performance of a reference predictor fitted without training samples.

The finite marginal contribution of data point $i$ to coalition $C\subseteq[N]\setminus\{i\}$ is
\begin{equation}\label{eqn:finite_marginal_target}
\Delta_i(C)=U(C\cup\{i\})-U(C).
\end{equation}
A valuation rule aggregates or approximates these quantities and returns a score vector $(\phi_1,\ldots,\phi_N)$.
In stochastic learning there are two related targets.
An \emph{algorithm-level} value compares independent counterfactual training runs on different coalitions.
A \emph{trajectory-conditioned} value asks for the first-order contribution to the particular stochastic training path that produced the model of interest.
NDDV estimates the second quantity from one forward--adjoint computation and then analyzes its discrepancy from Eq.~\eqref{eqn:finite_marginal_target}.
The two values are not equated by definition.

\subsection{Classical Data Valuation}
We next define the static reference quantities used in the comparison theorem.

\begin{definition}[Coalition Marginals and Symmetric Semi-Values]
\label{def:coalition_values}
Let $\Delta_i(C)$ be the finite add-one marginal in Eq.~\eqref{eqn:finite_marginal_target}.
For $k\in\{0,\ldots,N-1\}$, let $\mathcal C_k^{(-i)}=\{C\subseteq[N]\setminus\{i\}:|C|=k\}$ and define the size-$k$ average marginal
\[
\Delta_{k,i}(U)=\frac{1}{\binom{N-1}{k}}
\sum_{C\in\mathcal C_k^{(-i)}}\Delta_i(C).
\]
Given nonnegative weights $\omega_0,\ldots,\omega_{N-1}$ with $\sum_{k=0}^{N-1}\omega_k=1$, the associated symmetric semi-value is
\begin{equation}\label{eqn:symmetric_semivalue}
\phi_{\omega,i}(U)=\sum_{k=0}^{N-1}\omega_k\Delta_{k,i}(U).
\end{equation}
Leave-one-out uses $\omega_{N-1}=1$.
Shapley uses $\omega_k=1/N$.
Normalized Banzhaf uses $\omega_k=\binom{N-1}{k}/2^{N-1}$.
\end{definition}

Definition~\ref{def:coalition_values} uses finite counterfactual changes of the set utility.
They are the static reference values in this paper.
NDDV keeps marginal contribution as the target but estimates it from a local derivative of a fitted training trajectory.
The central theoretical task is not to relabel a sensitivity as a semi-value.
It is to bound the difference between the trajectory-conditioned sensitivity and the finite marginal contribution in Eq.~\eqref{eqn:finite_marginal_target}.

\section{Method}
\label{sec:NDDV}

NDDV is designed around one task: estimating pointwise marginal contributions from a single stochastic training trajectory.
The core computation consists of a controlled state process and its adjoint.
Each data point follows a state trajectory, the state trajectories are coupled through an empirical mean field, and the SMP converts a sample-level participation direction into a first-order change of the learning criterion.
The meta-weight map and the KAN--Mat\'ern terminal map specify the trajectory on which this derivative is evaluated.
They do not define additional value notions.

We use separate notation for the finite and trajectory-conditioned quantities.
The set function $U(C)$ is the validation utility after training on coalition $C$, and $\Delta_i(C)=U(C\cup\{i\})-U(C)$ is its finite marginal contribution.
A fitted NDDV run determines the trajectory context $\mathcal T=(\psi^K,\theta^K,\{X_{i,s}\}_{i,s},\{Y_{i,s}\}_{i,s},\{\mu_s\}_s,\{\Delta W_{i,s}\}_{i,s})$ used in the final scoring pass.
The raw trajectory sensitivity of data point $i$ is $s_i(\mathcal T)$, and the calibrated marginal-contribution score is $\phi_i^{\mathrm{NDDV}}$.
In population notation, $\mu_t$ denotes the aggregate state.
In the implemented Euler system, it is instantiated by the empirical aggregate $\mu_s=N^{-1}\sum_i v_iX_{i,s}$ on grid index $s$.
During the numerical backward sweep, the empirical aggregate produced by the forward pass is held fixed.
Superscripts such as $\mu_s^k$ index the outer iterations.

Algorithm~\ref{alg:pseudo_nddv} gives the computation used in the experiments.
Section~\ref{subsec:learning_stochastic_dynamic} first derives the score and then follows it through the population mean-field model, the discrete implementation, and the bridge to finite coalition marginal contribution.

\begin{algorithm}[tbp]
    \renewcommand{\algorithmicrequire}{\textbf{Input:}}
    \renewcommand{\algorithmicensure}{\textbf{Output:}}
    \caption{NDDV fitting and scoring procedure.} 
    \label{alg:pseudo_nddv}
    \begin{algorithmic}[1]  \footnotesize
        \Require Data $\mathcal D$, meta-data $\mathcal D'$, batch sizes $B,M_{\mathrm{meta}}$, Euler steps $S$, horizon $T$, iterations $K$, learning rates $\alpha,\beta$.
        \Ensure Sensitivities $\{s_i(\mathcal T)\}_{i=1}^N$ and scores $\{\phi_i^{\mathrm{NDDV}}\}_{i=1}^N$.
        \State Set $\Delta t=T/S$ and initialize $\psi^0=\{\psi_s^0\}_{s=0}^{S}$, $\theta^0$, and state variables.
        \For{$k=0$ {\bfseries to} $K-1$}
        \State Draw $\mathcal I_k\subset[N]$, $|\mathcal I_k|=B$, and $\mathcal M_k\subset\mathcal D'$, $|\mathcal M_k|=M_{\mathrm{meta}}$.
        \State Forward Euler: for $i\in\mathcal I_k$ and $s=0,\ldots,S-1$, record $X_{i,s}^k$, $\mu_s^k$, and $q_i^k=\Phi_i(X_{i,S}^{k},\mu_S^{k},\psi_S^{k})$.
        \State Form $\mu_s^k(\theta)=B^{-1}\sum_{i\in\mathcal I_k}\mathcal V_\theta(q_i^k)X_{i,s}^k$ and freeze $\mu_s^k=\mu_s^k(\theta^k)$.
        \State Set $Y_{i,S}^{k}=-\nabla_x\Phi_i(x,\mu_S^k,\psi_S^k)|_{x=X_{i,S}^{k}}$ and run the backward adjoint for $i\in\mathcal I_k$, $s=S-1,\ldots,0$:
        \begin{align*}
            Y_{i,s}^{k}=Y_{i,s+1}^{k}
            +\nabla_x\mathcal H_i(x,\mu_s^k,Y_{i,s+1}^{k},0,\psi_s^k)\big|_{x=X_{i,s}^{k}}\Delta t .
        \end{align*}
        \State Define $\hat\psi_s^k(\theta)=\psi_s^k+\alpha B^{-1}\sum_{i\in\mathcal I_k}\nabla_{\psi_s}\mathcal H_i(X_{i,s}^{k},\mu_s^k(\theta),Y_{i,s}^{k},0,\psi_s^k)$.
        \State Update $\theta^{k+1}=\theta^k-\beta M_{\mathrm{meta}}^{-1}\sum_{i\in\mathcal M_k}\nabla_\theta \ell_i(\hat\psi^k(\theta))|_{\theta=\theta^k}$.
        \State Update $\psi_s^{k+1}=\psi_s^k+\alpha B^{-1}\sum_{i\in\mathcal I_k}\nabla_{\psi_s}\mathcal H_i(X_{i,s}^{k},\mu_s^k,Y_{i,s}^{k},0,\psi_s^k)$ for $s=0,\ldots,S$.
        \EndFor
        \State Freeze $(\psi^K,\theta^K)$ and run one batched forward--adjoint scoring pass.
        \State Compute $s_i(\mathcal T)=-X_{i,S}^{\top}Y_{i,S}$, where $S$ indexes time $T$, with larger values denoting larger utility.
        \State Set $\phi_i^{\mathrm{NDDV}}=s_i(\mathcal T)-(N-1)^{-1}\sum_{j\ne i}s_j(\mathcal T)$ and rank data points by $\phi_i^{\mathrm{NDDV}}$.
    \end{algorithmic}
\end{algorithm}

\subsection{Trajectory-Conditioned Marginal Contribution}
\label{subsec:learning_stochastic_dynamic}

\begin{figure*}[t]
    \centering
    \includegraphics[width=\textwidth,height=0.72\textheight,keepaspectratio]{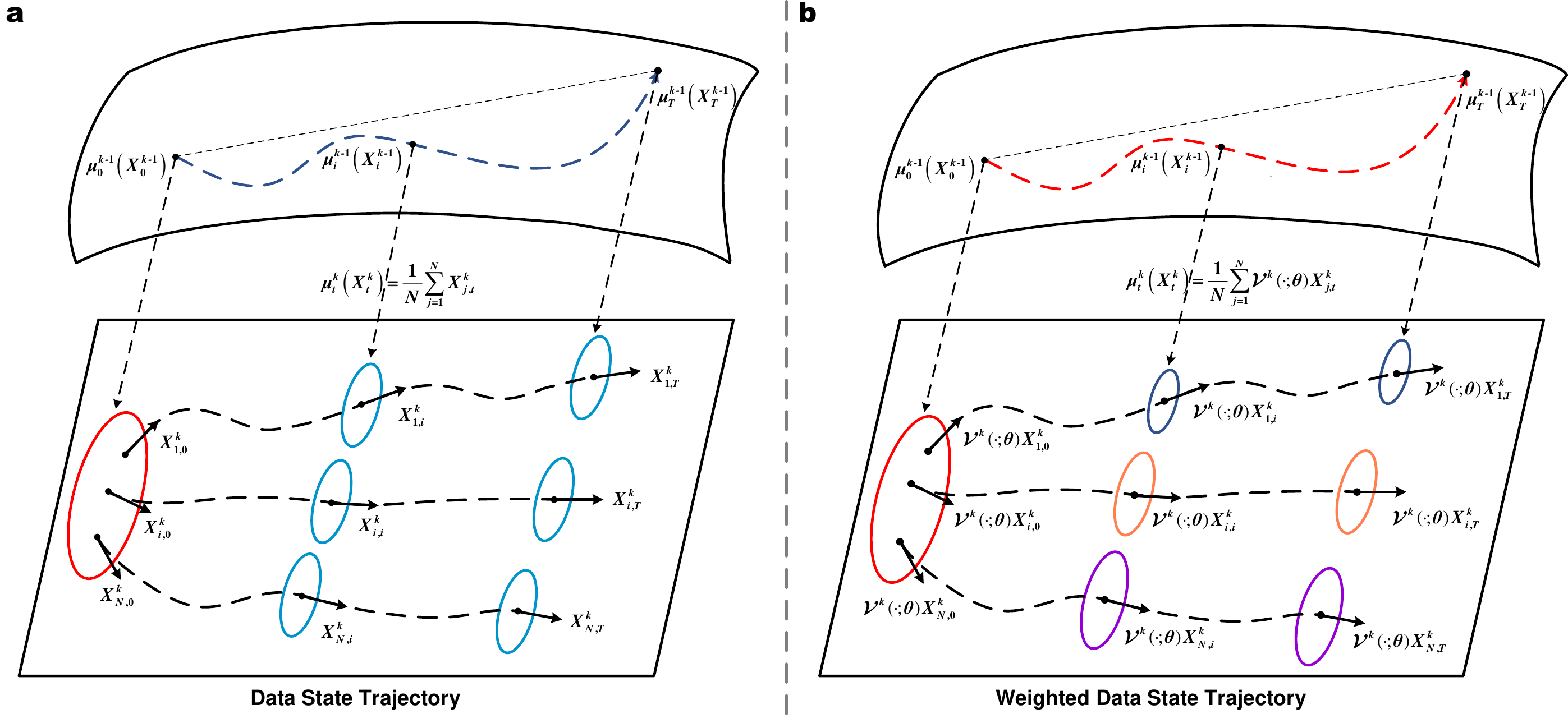}
    \vspace{-1em}
    \caption{Stochastic dynamic model for data valuation. a.
    Mean-field state trajectories. b.
    Meta-learned reweighting for heterogeneous data dynamics.}
    \label{fig:learn_sde}
\end{figure*}

We first formulate the one-run marginal-contribution problem as a sensitivity of controlled stochastic dynamics.
Increasing the participation of data point $i$ can change its initial representation, the drift that transports its state, or the running and terminal criteria.
The adjoint collects these first-order effects without resolving the forward system separately for each direction.
This forward system couples every data point's state through a shared empirical mean field, as illustrated in Figure~\ref{fig:learn_sde}~\citep{casert2024learning,gao2024learning}.

The derivation distinguishes three levels.
The reference object is a weighted mean-field control problem with aggregate $\mu_t\in\mathbb R^d$.
The NDDV surrogate freezes the empirical aggregate generated by the forward pass before computing sample-wise adjoints.
Finally, the implementation differentiates the sampled Euler trajectory pathwise.
This backward recursion is exact for the discretized frozen-aggregate objective.
Conditional projection and time refinement connect it to the adapted continuous-time SMP adjoint.

Let $\psi:[0,T]\to\Psi\subset\mathbb R^p$ be the shared control path, $X_{i,t}\in\mathbb R^d$ the state of data point $i$, and $\mathbf X_t=(X_{1,t},\ldots,X_{N,t})$ the stacked state.
The drift is $b:\mathbb R^d\times\mathbb R^d\times\Psi\to\mathbb R^d$.
Here $W_{i,t}\in\mathbb R^m$ is a Wiener process, $\Sigma\in\mathbb R^{d\times m}$ is the diffusion matrix, and $Z_{i,t}\in\mathbb R^{d\times m}$ is the martingale integrand.
In the linear--quadratic specialization, either $p=d$ or a fixed linear map embeds $\psi_t$ in the state space.
The reference expected cost is
\begin{equation}\label{eqn:cost_function}
\mathcal{L}(\psi)=\mathbb{E}\!\left[\int_0^{T} R(\mathbf X_t,\mu_t,\psi_t)\,\mathrm{d}t + \Phi(\mathbf X_T,\mu_T,\psi_T)\right].
\end{equation}
Here $R=N^{-1}\sum_i R_i$ and $\Phi=N^{-1}\sum_i \Phi_i$ denote the averaged running and terminal costs.
Each reference state satisfies
\begin{equation}\label{eqn:controlled_SDE}
\mathrm{d}X_{i,t}=b(X_{i,t},\mu_t,\psi_t)\,\mathrm{d}t+\Sigma\,\mathrm{d}W_{i,t}, \qquad X_{i,0}=x_i.
\end{equation}
The SMP pairs the forward state with a backward co-state that carries sensitivity of the objective.
In a full McKean--Vlasov problem, this adjoint also contains derivatives with respect to the population law~\citep{carmona2015forward,carmona2013control}.
We instead condition on the empirical path obtained in the forward pass and hold it fixed while solving the sample-wise backward equations.
The resulting surrogate Hamiltonian is
\begin{equation}\label{eqn:ham}
\mathcal{H}_i(X_{i,t},\mu_t,Y_{i,t},Z_{i,t},\psi_t)
= b(X_{i,t},\mu_t,\psi_t)\!\cdot\!Y_{i,t}
  +\mathrm{tr}(\Sigma^\top Z_{i,t})
  - R_i(X_{i,t},\mu_t,\psi_t).
\end{equation}
The surrogate adjoint equation is
\begin{equation}\label{eqn:adjoint_eq}
\begin{aligned}
\mathrm{d}Y_{i,t}
&=-\nabla_x\mathcal{H}_i(x,\mu_t,Y_{i,t},Z_{i,t},\psi_t)\big|_{x=X_{i,t}}\,\mathrm{d}t
  + Z_{i,t}\,\mathrm{d}W_{i,t},\\
Y_{i,T}&=-\nabla_x\Phi_i(x,\mu_T,\psi_T)\big|_{x=X_{i,T}}.
\end{aligned}
\end{equation}

\begin{assumption}[Convex-control sufficiency]
\label{assump:convexity}
For almost every $t$, the control set $\Psi$ is convex, $R_i(x,\mu,\psi)$ is jointly convex in $(x,\mu,\psi)$, $\Phi_i(x,\mu,\psi)$ is convex in $(x,\mu)$, and $b(x,\mu,\psi)$ is affine in $(x,\mu,\psi)$.
These conditions are used only when Eq.~\eqref{eqn:max_ham} is interpreted as sufficient for optimality.
The sensitivity results below require differentiability and moment bounds but not global convexity.
\end{assumption}

With the maximization convention, an admissible shared control obeys the aggregate Pontryagin stationarity condition\footnote{We follow the maximization convention of~\citet{li2018optimal}, for which $Y_T=-\nabla_x\Phi(X_T)$.
Replacing the terminal cost $h$ in~\citet[Theorem~3.1]{yong1999stochastic} by $-\Phi$ gives the equivalent standard convention.}
\begin{equation}\label{eqn:max_ham}
\psi_t^*\in\arg\max_{\psi\in\Psi}
\frac{1}{N}\sum_{i=1}^N
\mathcal H_i(X_{i,t}^*,\mu_t,Y_{i,t}^*,Z_{i,t}^*,\psi),
\end{equation}
for almost every $t\in[0,T]$.
In practice, mini-batches estimate the population average.
Under Assumption~\ref{assump:convexity}, convexity of the costs and affinity of the drift make the condition sufficient.
Outside that setting, Eq.~\eqref{eqn:max_ham} is a first-order stationarity condition.

\begin{assumption}[Regularity of the frozen-aggregate model]
\label{assump:adjoint_regular}
Along any bounded frozen aggregate path, $b(\cdot,\mu_t,\psi_t)$, $R_i(\cdot,\mu_t,\psi_t)$, and $\Phi_i(\cdot,\mu_T,\psi_T)$ are twice continuously differentiable in the state.
Their first derivatives are globally Lipschitz and their second derivatives are uniformly bounded.
The diffusion is additive, and the state and terminal derivatives have finite second moments.
\end{assumption}

\begin{theorem}[Adjoint representation]
\label{thm:general_adjoint_perturbation}
Fix a bounded aggregate path $\mu$ and an admissible control $\psi$.
Let a scalar marginal-contribution parameter $\epsilon$ change the initial state, drift, running cost, and terminal cost according to
\begin{align}
X_{i,0}^{\epsilon}&=x_i+\epsilon\xi_i,\nonumber\\
\mathrm dX_{i,t}^{\epsilon}
&=\left[b(X_{i,t}^{\epsilon},\mu_t,\psi_t)
+\epsilon h_{i,t}(X_{i,t}^{\epsilon})\right]\mathrm dt
+\Sigma\,\mathrm dW_{i,t},
\label{eqn:general_perturbed_state}\\
\mathcal J_i(\epsilon)
&=\mathbb E\!\left[
\int_0^T\!\left(
R_i(X_{i,t}^{\epsilon},\mu_t,\psi_t)
+\epsilon r_{i,t}(X_{i,t}^{\epsilon})
\right)\mathrm dt
+\Phi_i(X_{i,T}^{\epsilon},\mu_T,\psi_T)
+\epsilon q_i(X_{i,T}^{\epsilon})
\right],
\label{eqn:general_perturbed_objective}
\end{align}
Suppose $\xi_i$ is square integrable and $h_{i,t}$, $r_{i,t}$, and $q_i$ satisfy the regularity needed to interchange differentiation and expectation.
Let $(Y_i,Z_i)$ solve the frozen-aggregate adjoint equation~\eqref{eqn:adjoint_eq}.
Then
\begin{equation}
\label{eqn:general_adjoint_representation}
\mathcal J_i'(0)
=
\mathbb E\!\left[
q_i(X_{i,T})-Y_{i,0}\!\cdot\!\xi_i
+\int_0^T
\left(r_{i,t}(X_{i,t})-Y_{i,t}\!\cdot\!h_{i,t}(X_{i,t})\right)\mathrm dt
\right].
\end{equation}
Thus marginal effects entering along the training dynamics are valued by the adjoint path.
If the only sensitivity direction is the terminal rescaling $\Phi_i^{\epsilon}(x,\mu_T,\psi_T)=\Phi_i((1+\epsilon)x,\mu_T,\psi_T)$, then $q_i(x)=x\cdot\nabla_x\Phi_i(x,\mu_T,\psi_T)$ and Eq.~\eqref{eqn:general_adjoint_representation} reduces to the terminal NDDV sensitivity in Eq.~\eqref{eqn:utility}.
\end{theorem}

\noindent\textbf{Frozen-Aggregate Well-Posedness and Stability.}
For $b(X_{i,t},\mu_t,\psi_t)=a(\mu_t-X_{i,t})+\psi_t$, the drift is affine in the state and control and uniformly Lipschitz along any bounded aggregate path.
Standard linear--quadratic FBSDE results give a unique forward state and a unique frozen-aggregate adjoint for square-integrable terminal data~\citep[Theorem~2.1]{carmona2013control}.

To quantify the approximation, let $(X_{i,t}^\star,Y_{i,t}^\star,Z_{i,t}^\star)$ denote a population mean-field state and adjoint.
We write its adjoint driver as the frozen driver plus the law-derivative contribution $\mathfrak M_{i,t}$:
\begin{equation}\label{eqn:ideal_adjoint_with_measure_term}
\begin{aligned}
\mathrm dY_{i,t}^\star
&=-\left[
\nabla_x\mathcal H_i(X_{i,t}^\star,\mu_t^\star,Y_{i,t}^\star,Z_{i,t}^\star,\psi_t)
+\mathfrak M_{i,t}
\right]\mathrm dt
+Z_{i,t}^\star\,\mathrm dW_{i,t},\\
Y_{i,T}^\star
&=-\nabla_x\Phi_i(X_{i,T}^\star,\mu_T^\star,\psi_T)
-\mathfrak M_{i,T}^{\Phi}.
\end{aligned}
\end{equation}
Here $\mathfrak M_{i,t}$ and $\mathfrak M_{i,T}^{\Phi}$ collect the derivatives of the drift and costs with respect to the population law, including the independent-copy terms that appear in the McKean--Vlasov maximum principle.
Let $(\bar X_{i,t},\bar Y_{i,t},\bar Z_{i,t})$ solve the frozen-aggregate system along $\bar\mu_t$.
Define
\[
\eta_\mu
=\left\|\sup_{0\le t\le T}\|\mu_t^\star-\bar\mu_t\|\right\|_{L^2},
\qquad
\Gamma_{\mathrm{law}}
=\left(
\mathbb E\|\mathfrak M_{i,T}^{\Phi}\|^2
+\mathbb E\int_0^T\|\mathfrak M_{i,t}\|^2\,\mathrm dt
\right)^{1/2}.
\]

\begin{proposition}[Frozen-Aggregate Stability]
\label{prop:frozen_mean_field_stability}
Under Assumption~\ref{assump:adjoint_regular}, suppose $b$, $\nabla_xR_i$, and $\nabla_x\Phi_i$ are also Lipschitz in the aggregate variable.
There is a constant $C_{\mathrm{MF}}$, depending only on the Lipschitz constants, $T$, and second-moment bounds, such that
\begin{align}
&\|X_i^\star-\bar X_i\|_{\mathcal S^2}
+\|Y_i^\star-\bar Y_i\|_{\mathcal S^2}
+\|Z_i^\star-\bar Z_i\|_{\mathcal H^2}
\le C_{\mathrm{MF}}(\eta_\mu+\Gamma_{\mathrm{law}}),
\label{eqn:frozen_mean_field_stability}\\
&\mathbb E\left|
X_{i,T}^\star\!\cdot Y_{i,T}^\star
-\bar X_{i,T}\!\cdot\bar Y_{i,T}
\right|
\le C_{\mathrm{score}}(\eta_\mu+\Gamma_{\mathrm{law}}).
\label{eqn:frozen_score_stability}
\end{align}
Here $\mathcal S^2$ is the $L^2$ norm of the path supremum and $\mathcal H^2$ is the $L^2$ norm in time.
The display itself does not imply a universal empirical-aggregate rate for learned, dependent weights.
If one separately verifies or assumes $\eta_\mu\le C_\mu N^{-1/2}$ and the law-interaction strength obeys $\Gamma_{\mathrm{law}}\le\gamma_{\mathrm{law}}$, then the terminal sensitivity error is $O(N^{-1/2}+\gamma_{\mathrm{law}})$.
\end{proposition}

\noindent\textbf{Pathwise Discrete Adjoint.}
The adapted BSDE in Eq.~\eqref{eqn:adjoint_eq} is the natural continuous-time object, whereas the implementation differentiates a sampled Euler trajectory after the noise increments and the empirical aggregate have been realized.
The next result identifies this computation exactly.

Let $0=t_0<\cdots<t_S=T$, $\Delta t=T/S$, and fix the aggregate values $\bar\mu_s$, controls $\psi_s$, and noise increments $\Delta W_{i,s}$.
Define
\begin{equation}\label{eqn:discrete_state_map}
X_{i,s+1}^{\Delta}=F_{i,s}(X_{i,s}^{\Delta})
:=X_{i,s}^{\Delta}+b(X_{i,s}^{\Delta},\bar\mu_s,\psi_s)\Delta t+\Sigma\Delta W_{i,s}.
\end{equation}
For a state value $x$ at step $s$, let $X_{i,\ell}^{x,\Delta}$ denote the continuation generated by Eq.~\eqref{eqn:discrete_state_map}, and define the sampled tail objective
\begin{equation}\label{eqn:discrete_tail_objective}
J_{i,s}^{\Delta}(x)
=
\sum_{\ell=s}^{S-1}R_i(X_{i,\ell}^{x,\Delta},\bar\mu_\ell,\psi_\ell)\Delta t
+\Phi_i(X_{i,S}^{x,\Delta},\bar\mu_S,\psi_S).
\end{equation}

\begin{theorem}[Exact Discrete-Adjoint Identity]
\label{thm:discrete_adjoint_identity}
Under Assumption~\ref{assump:adjoint_regular}, the backward recursion used in Algorithm~\ref{alg:pseudo_nddv} satisfies
\begin{equation}\label{eqn:discrete_adjoint_identity}
Y_{i,s}^{\Delta}
=-\nabla_xJ_{i,s}^{\Delta}(X_{i,s}^{\Delta}),
\qquad s=0,\ldots,S.
\end{equation}
Equivalently, for any direction $\xi\in\mathbb R^d$,
\begin{equation}\label{eqn:discrete_directional_identity}
\left.\frac{\mathrm d}{\mathrm d\epsilon}
J_{i,s}^{\Delta}(X_{i,s}^{\Delta}+\epsilon\xi)
\right|_{\epsilon=0}
=-Y_{i,s}^{\Delta}\!\cdot\xi.
\end{equation}
Thus the numerical backward sweep is exact for the sampled, frozen-aggregate Euler objective.
It is not an additional heuristic once that discrete objective has been fixed.
\end{theorem}

\begin{proposition}[Conditional projection to the SMP co-state]
\label{prop:conditional_projection}
Under Assumption~\ref{assump:adjoint_regular}, let $\bar Y_{i,t}$ be the pathwise backward process along the continuous frozen-aggregate trajectory,
\begin{equation}\label{eqn:pathwise_adjoint}
\bar Y_{i,t}
=-\nabla_x\Phi_i(X_{i,T},\mu_T,\psi_T)
+\int_t^T\nabla_x\mathcal H_i(X_{i,s},\mu_s,\bar Y_{i,s},0,\psi_s)\,\mathrm ds.
\end{equation}
Although $\bar Y_{i,t}$ need not be adapted, its conditional projection
\begin{equation}\label{eqn:conditional_projection}
Y_{i,t}=\mathbb E[\bar Y_{i,t}\mid\mathcal F_t]
\end{equation}
solves Eq.~\eqref{eqn:adjoint_eq} for a unique martingale integrand $Z_{i,t}$.
Consequently, the $Z$ term records the martingale representation needed to adapt the pathwise sensitivity.
The implementation computes a sampled pathwise adjoint and stochastic averaging estimates its expected effect.
\end{proposition}

\begin{corollary}[Terminal-Score Numerical Error]
\label{cor:numerical_score_error}
Let $h_i(x)=x\cdot\nabla_x\Phi_i(x,\mu_T,\psi_T)$ be $L_{h,i}$-Lipschitz, and suppose Euler--Maruyama satisfies $\|X_{i,T}-X_{i,S}^{\Delta}\|_{L^2}\le C_{X,i}\Delta t^{1/2}$.
For $M_{\mathrm{traj}}$ independent sampled trajectories, define
\[
\widehat h_i^{M_{\mathrm{traj}},\Delta}
=\frac1{M_{\mathrm{traj}}}\sum_{m=1}^{M_{\mathrm{traj}}} h_i(X_{i,S}^{\Delta,(m)}),
\qquad
\sigma_{h,i}^2=\sup_{\Delta t}\operatorname{Var}(h_i(X_{i,S}^{\Delta})).
\]
Then
\begin{equation}\label{eqn:numerical_score_error}
\left\|\widehat h_i^{M_{\mathrm{traj}},\Delta}-\mathbb E[h_i(X_{i,T})]\right\|_{L^2}
\le
L_{h,i}C_{X,i}\Delta t^{1/2}
+\frac{\sigma_{h,i}}{\sqrt{M_{\mathrm{traj}}}}.
\end{equation}
The first term is a time-discretization error and the second is a sampling error.
\end{corollary}

\noindent\textbf{Mass-Preserving Sensitivity.}
For data point $i$, define the oriented terminal trajectory sensitivity
\begin{equation}\label{eqn:utility}
s_i(\mathcal T)
:= X_{i,T}\!\cdot\!\nabla_x\Phi_i(x,\mu_T,\psi_T)\big|_{x=X_{i,T}}
= -X_{i,T}\!\cdot\!Y_{i,T},
\end{equation}
where $\mathcal T$ denotes the fitted trajectory context.
Equation~\eqref{eqn:utility} is the terminal radial specialization of Theorem~\ref{thm:general_adjoint_perturbation}.
It has the algebraic form of Gradient$\times$Input~\citep{shrikumar2017learning,ancona2019explaining}, but the state is generated by the coupled stochastic-control system.
When $\Phi_i$ is implemented as a loss, we apply one global sign so that larger scores have the same orientation as the higher-is-better set utility $U(C)$.
This orientation is absorbed into the notation.

A coordinate derivative changes the participation of one data point without controlling the total data mass.
For a relative marginal contribution, let $e_i$ be the $i$th coordinate vector and define the mass-preserving direction
\begin{equation}\label{eqn:mass_preserving_direction}
d_i=e_i-\frac{1}{N-1}\sum_{j\ne i}e_j.
\end{equation}
For the fixed trajectory context, consider the oriented terminal functional
\[
\mathcal G_{\mathcal T}(\lambda)
=\sum_{j=1}^N
\Phi_j(\lambda_jX_{j,T},\mu_T,\psi_T),
\qquad \lambda\in\mathbb R^N,
\]
with the same global orientation convention as Eq.~\eqref{eqn:utility}.
The final NDDV score is
\begin{equation}\label{eqn:dynamic_dv}
\phi_i^{\mathrm{NDDV}}
= s_i(\mathcal T)
- \frac{1}{N-1}\sum_{j\ne i}s_j(\mathcal T).
\end{equation}

\begin{proposition}[Mass-preserving calibration]
\label{prop:mass_preserving_calibration}
If each terminal map is differentiable, then
\begin{equation}\label{eqn:mass_preserving_derivative}
D\mathcal G_{\mathcal T}(\mathbf 1)[d_i]
=\phi_i^{\mathrm{NDDV}}.
\end{equation}
Thus $\phi_i^{\mathrm{NDDV}}$ is the first-order change obtained by increasing the participation of data point $i$ while redistributing the same total participation uniformly over the remaining data points.
\end{proposition}

\begin{proof}
The chain rule gives $\partial_{\lambda_j}\mathcal G_{\mathcal T}(\mathbf 1)=s_j(\mathcal T)$.
Taking the directional derivative along Eq.~\eqref{eqn:mass_preserving_direction} yields Eq.~\eqref{eqn:dynamic_dv}.
\end{proof}

A trained run fixes $\mathcal T$.
The meta dataset $\mathcal D'$ updates only $\mathcal V_\theta$, and $\psi_t$ denotes the shared parameters at time $t$.
Because
\[
\phi_i^{\mathrm{NDDV}}-\phi_j^{\mathrm{NDDV}}
=\frac{N}{N-1}\bigl(s_i(\mathcal T)-s_j(\mathcal T)\bigr),
\]
the calibration preserves the pairwise ordering of the raw sensitivities.

\begin{proposition}[First-order terminal sensitivity]
\label{prop:first_order_terminal_sensitivity}
If $\Phi_i$ is differentiable in its state argument, then $s_i(\mathcal T)$ is the directional derivative of the terminal scalar under the radial variation $X_{i,T}\mapsto(1+\eta)X_{i,T}$.
This is the terminal specialization of Theorem~\ref{thm:general_adjoint_perturbation}.
\end{proposition}

\begin{proof}
Let $F_i(\eta)=\Phi_i((1+\eta)X_{i,T},\mu_T,\psi_T)$.
The chain rule and the terminal condition in Eq.~\eqref{eqn:adjoint_eq} give
\[
F_i'(0)=X_{i,T}\cdot\nabla_x\Phi_i(X_{i,T},\mu_T,\psi_T) =-X_{i,T}\cdot Y_{i,T}=s_i(\mathcal T).
\]
\end{proof}

\begin{definition}[Continuous marginal-contribution path]
\label{def:continuous_inclusion}
For $C\subseteq[N]\setminus\{i\}$, let $U_{C,i}(\lambda)$ be the validation utility produced by the same learning rule when data point $i$ enters with continuous sample weight $\lambda\in[0,1]$, with $U_{C,i}(0)=U(C)$ and $U_{C,i}(1)=U(C\cup\{i\})$.
Whenever the derivative exists, write
\[
m_i(C,\lambda)=\partial_\lambda U_{C,i}(\lambda)
\]
for the local marginal sensitivity.
\end{definition}

\begin{proposition}[Finite marginal as an integral of local sensitivities]
\label{prop:continuous_marginal_identity}
If $U_{C,i}$ is absolutely continuous, then
\begin{equation}\label{eqn:continuous_marginal_identity}
U(C\cup\{i\})-U(C)=\int_0^1m_i(C,\lambda)\,\mathrm d\lambda.
\end{equation}
If $m_i(C,\cdot)$ is $L_{i,C}$-Lipschitz, then for any $\lambda_0\in[0,1]$,
\begin{equation}\label{eqn:local_to_finite_marginal}
\left|
U(C\cup\{i\})-U(C)-m_i(C,\lambda_0)
\right|
\le
\frac{L_{i,C}}{2}\bigl(\lambda_0^2+(1-\lambda_0)^2\bigr).
\end{equation}
The midpoint $\lambda_0=1/2$ gives the smallest worst-case constant, $L_{i,C}/4$.
\end{proposition}

\begin{assumption}[Local terminal calibration]
\label{assump:local_terminal_calibration}
For the coalition $C$, data point $i$, and interpolation point $\lambda_0$ under consideration, the inclusion-path utility is differentiable near $\lambda_0$ and admits the local decomposition
\begin{equation}\label{eqn:utility_calibration_decomposition}
U_{C,i}(\lambda)=\Gamma_{C,i}(\lambda)+g_i(X_{i,T}^{C,\lambda}),
\end{equation}
where $g_i$ is the oriented terminal scalar and $\Gamma_{C,i}$ collects the part of the validation utility not represented by that scalar.
The functions $\Gamma_{C,i}$ and $g_i$ are differentiable at the values used below.
\end{assumption}

\begin{proposition}[Local adjoint calibration]
\label{prop:local_adjoint_calibration}
Fix $C$, $i$, and $\lambda_0$, and suppose Assumption~\ref{assump:local_terminal_calibration} holds.
If $X_{i,T}^{C,\lambda_0}\neq0$, define the orthogonal radial--residual decomposition
\begin{equation}\label{eqn:inclusion_tangent_decomposition}
\partial_\lambda X_{i,T}^{C,\lambda_0}
=\rho_i(C,\lambda_0)X_{i,T}^{C,\lambda_0}+r_i(C,\lambda_0),
\qquad
\rho_i(C,\lambda_0)
=\frac{\left\langle\partial_\lambda X_{i,T}^{C,\lambda_0},X_{i,T}^{C,\lambda_0}\right\rangle}
{\|X_{i,T}^{C,\lambda_0}\|^2},
\end{equation}
so that $r_i(C,\lambda_0)\perp X_{i,T}^{C,\lambda_0}$.
If $X_{i,T}^{C,\lambda_0}=0$, set $\rho_i(C,\lambda_0)=0$ and $r_i(C,\lambda_0)=\partial_\lambda X_{i,T}^{C,\lambda_0}$.
Let
\[
u_i^{C,\lambda_0} =X_{i,T}^{C,\lambda_0}\cdot\nabla g_i(X_{i,T}^{C,\lambda_0}).
\]
Then
\begin{align}
\left|m_i(C,\lambda_0)-u_i^{C,\lambda_0}\right|
&\le
\left|\Gamma_{C,i}'(\lambda_0)\right|
+\left\|\nabla g_i(X_{i,T}^{C,\lambda_0})\right\|
 \left\|r_i(C,\lambda_0)\right\| \nonumber\\
&\quad
+\left|\rho_i(C,\lambda_0)-1\right|
 \left|u_i^{C,\lambda_0}\right|.
\label{eqn:local_adjoint_calibration}
\end{align}
If $\varepsilon_i^{\mathrm{ctx}}(C,\lambda_0) =|s_i(\mathcal T)-u_i^{C,\lambda_0}|$, then the right-hand side of Eq.~\eqref{eqn:local_adjoint_calibration} plus $\varepsilon_i^{\mathrm{ctx}}$ bounds $|m_i(C,\lambda_0)-s_i(\mathcal T)|$.
The context term can be split further into frozen-mean-field, fitting, discretization, and sampling components.
Corollary~\ref{cor:numerical_score_error} controls the last two under its assumptions.
\end{proposition}

\begin{theorem}[Local-to-finite marginal bridge]
\label{thm:continuous_inclusion_bridge}
Under Proposition~\ref{prop:continuous_marginal_identity}, Assumption~\ref{assump:local_terminal_calibration}, and Proposition~\ref{prop:local_adjoint_calibration}, define
\begin{align}
B_i(C;\lambda_0)
&=
\frac{L_{i,C}}{2}\bigl(\lambda_0^2+(1-\lambda_0)^2\bigr)
+\left|\Gamma_{C,i}'(\lambda_0)\right| 
+\left\|\nabla g_i\right\|\left\|r_i\right\| \nonumber\\
&\quad
+\left|\rho_i-1\right|\left|u_i^{C,\lambda_0}\right|
+\varepsilon_i^{\mathrm{ctx}}(C,\lambda_0),
\label{eqn:coalition_bridge_budget}
\end{align}
where the quantities on the second line are evaluated at $(C,\lambda_0)$.
Then
\begin{equation}\label{eqn:coalition_bridge}
\left|
s_i(\mathcal T)-\bigl[U(C\cup\{i\})-U(C)\bigr]
\right|
\le B_i(C;\lambda_0).
\end{equation}
Thus $B_i$ separates the gap into inclusion-path curvature, terminal-direction mismatch, scale mismatch, validation-to-terminal calibration, and model-to-implementation error.
\end{theorem}

\begin{theorem}[Pair-specific comparison with symmetric semi-values]
\label{thm:ranking_consistency}
Let $p_k^\omega=\omega_k/\binom{N-1}{k}$ for $0\le k\le N-1$ and $p_N^\omega=0$.
For $A\subseteq[N]\setminus\{i,j\}$, define
\begin{equation}\label{eqn:semivalue_pair_weights}
c_A^\omega=p_{|A|}^\omega+p_{|A|+1}^\omega.
\end{equation}
Then $c_A^\omega\ge0$ and $\sum_{A\subseteq[N]\setminus\{i,j\}}c_A^\omega=1$.
All sums over $A$ in this theorem use this domain.
Let
\[
\Delta_i(A)=U(A\cup\{i\})-U(A),
\qquad
e_i(A)=\left|s_i(\mathcal T)-\Delta_i(A)\right|,
\]
and define the pair-specific weighted error
\begin{equation}\label{eqn:pair_specific_error}
\mathcal E_{\omega,ij}
=\sum_{A\subseteq[N]\setminus\{i,j\}}
c_A^\omega\bigl(e_i(A)+e_j(A)\bigr).
\end{equation}
For the symmetric semi-value associated with $\omega$,
\begin{equation}\label{eqn:ranking_bound}
\left|
\frac{N-1}{N}\bigl(\phi_i^{\mathrm{NDDV}}-\phi_j^{\mathrm{NDDV}}\bigr)
-\bigl(\phi_{\omega,i}(U)-\phi_{\omega,j}(U)\bigr)
\right|
\le\mathcal E_{\omega,ij}.
\end{equation}
Moreover, Theorem~\ref{thm:continuous_inclusion_bridge} gives the deterministic upper bound
\begin{equation}\label{eqn:pair_specific_theoretical_budget}
\mathcal E_{\omega,ij}
\le
\sum_Ac_A^\omega
\bigl(B_i(A;\lambda_0)+B_j(A;\lambda_0)\bigr).
\end{equation}
If the absolute value of either the semi-value gap or the rescaled NDDV gap exceeds $\mathcal E_{\omega,ij}$, the two methods rank $i$ and $j$ in the same order.
The result includes LOO, Shapley, and normalized Banzhaf through their respective choices of $\omega$.
\end{theorem}

\begin{corollary}[Finite-sample pairwise certificate]
\label{cor:sampled_pairwise_certificate}
Condition on a trained model.
Let $\mathbb P_{\omega,ij}$ assign mass $c_A^\omega$ to each admissible coalition $A$.
Sample $A_1,\ldots,A_{M_{\mathrm{coal}}}$ independently from this distribution.
Let
\[
Z_m=e_i(A_m)+e_j(A_m),
\qquad
\widehat{\mathcal E}_{\omega,ij}=\frac1{M_{\mathrm{coal}}}\sum_{m=1}^{M_{\mathrm{coal}}}Z_m,
\]
and assume $0\le Z_m\le B_{ij}^{\max}$ almost surely.
Then, with probability at least $1-\alpha$ over the sampled coalitions,
\begin{equation}\label{eqn:sampled_pairwise_certificate}
\mathcal E_{\omega,ij}
\le
\widehat{\mathcal E}_{\omega,ij}
+B_{ij}^{\max}\sqrt{\frac{\log(1/\alpha)}{2M_{\mathrm{coal}}}}.
\end{equation}
Consequently, if the absolute rescaled NDDV gap exceeds the right-hand side, its sign agrees with the true symmetric semi-value gap with probability at least $1-\alpha$.
A union bound replaces $\alpha$ by $\alpha/P$ when $P$ pre-specified pairs are checked simultaneously.
\end{corollary}

Proofs of the general adjoint representation, model-to-frozen stability bound, exact discrete-adjoint identity, and semi-value comparison results appear in Appendices~\ref{app:data_state_utility} and~\ref{app:ranking_consistency}.
Unlike a global supremum bound, $\mathcal E_{\omega,ij}$ is pair-specific and weighted by the coalition distribution of the selected semi-value.

\noindent\textbf{Numerical Stability.}
We differentiate the sampled Euler trajectory by automatic differentiation and clip the gradient norm at $1.0$.
The activations are Lipschitz bounded, and the state Jacobian of the LQ drift has eigenvalues in $[-a,0]$.
Across five seeds on the six benchmarks, the computed adjoints satisfy $\|Y_{i,0}\|\le5\|Y_{i,T}\|$.

\noindent\textbf{Time-Resolved Scores.}
The same state--adjoint pairing can be evaluated at intermediate layers or times.
Replacing $(X_{i,T},Y_{i,T})$ with $(X_{i,t}^k,Y_{i,t}^k)$ at outer iteration $k$, we define
\begin{equation}\label{eqn:layer_epoch_phi}
\phi^k_{i,t} 
= -X^k_{i,t}\!\cdot\!Y^k_{i,t} 
+ \sum_{j\ne i}\frac{X^k_{j,t}\!\cdot\!Y^k_{j,t}}{N-1}.
\end{equation}
Equation~\eqref{eqn:layer_epoch_phi} is used to trace the fitted sensitivity through depth and training iterations.
It does not define a separate valuation target.
Let $q_i^k=\Phi_i(X^k_{i,T},\mu_T^k,\psi_T^k)$ denote the terminal signal.
The state and adjoint paths at a fixed outer iteration are
\[
\begin{tikzcd}[column sep=small,row sep=large]
x_i=X^k_{i,0} \arrow[r] 
& X^k_{i,1} \arrow[r] 
& \cdots \arrow[r]
& X^k_{i,t} \arrow[r] 
& \cdots \arrow[r]
& X^k_{i,T} \arrow[d, "q_i^k"]\\
Y^k_{i,0} 
& Y^k_{i,1} \arrow[l] 
& \cdots \arrow[l]
& Y^k_{i,t} \arrow[l] 
& \cdots \arrow[l]
& Y^k_{i,T} \arrow[l]
\end{tikzcd}
\]
The epoch-wise path at the terminal layer is shown as follows.
\[
\begin{tikzcd}[column sep=small,row sep=large]
X^0_{i,T} \arrow[r] 
& X^1_{i,T} \arrow[r] 
& \cdots \arrow[r]
& X^k_{i,T} \arrow[r] 
& \cdots \arrow[r]
& X^K_{i,T} \arrow[d, "q_i^K"]\\
Y^0_{i,T} 
& Y^1_{i,T} \arrow[l] 
& \cdots \arrow[l]
& Y^k_{i,T} \arrow[l] 
& \cdots \arrow[l]
& Y^K_{i,T} \arrow[l]
\end{tikzcd}
\]
The states evolve from left to right, while the adjoints propagate backward from the terminal condition.
Together, the two diagrams show where the estimated sample sensitivity changes across depth and training.
They are obtained from the fitted coupled system~\citep{serban_cvodes_2005,jorgensen_2007} and require no enumeration of data coalitions.

\subsection{Mean-Field Coupling, Reweighting, and Relative-State Stability}
\label{subsec:mean_field_stability}

The mean field specifies how samples interact in the trajectory on which marginal contribution is evaluated.
A learned scalar $v_i$ controls the contribution of $X_{i,t}$ to the empirical aggregate, but it is not multiplied into the score in Eq.~\eqref{eqn:dynamic_dv}.
Reweighting changes the fitted trajectory and terminal map rather than introducing a second valuation rule.
The Hamiltonian and adjoint retain the form of Section~\ref{subsec:learning_stochastic_dynamic}.

The relative-state analysis below clarifies what this coupling can and cannot do.
In the linear--quadratic model, the common aggregate cancels from pairwise state differences, so the interaction coefficient controls contraction while the learned weights act indirectly through the common path and terminal map.
Class-conditional score and detection gaps are used later as auxiliary diagnostics.
The six OpenDataVal benchmarks contain no protected attributes.
Appendix~\ref{app:protected_attribute_validation} gives a limited Adult check in which sex is available.

\noindent\textbf{Meta-Learned Sample Weights.}
We set $v_i=\mathcal V_\theta(q_i)$, where $q_i=\Phi_i(X_{i,S},\mu_S,\psi_S)$ is the terminal-grid loss signal from the current forward pass.
We freeze $q_i$ before evaluating the weight, which avoids an implicit fixed point among $q_i$, $v_i$, and $\mu_t$.
This construction follows sample-reweighting ideas related to Stackelberg formulations~\citep{bensoussan2017linear}.
The weights enter the dynamics through $\mu_s(\theta)=N^{-1}\sum_i v_iX_{i,s}$ on the Euler grid, and the resulting discrete control objective is
\begin{equation}\label{eqn:meta_objective}
L(\psi,\theta)=\frac{1}{N}\sum_{i=1}^{N}\!\left[\sum_{s=0}^{S-1}R_i(X_{i,s},\mu_s(\theta),\psi_s)\Delta t+\Phi_i(X_{i,S},\mu_S(\theta),\psi_S)\right].
\end{equation}
A ReLU--sigmoid network constrains $\mathcal V_\theta$ to $[0,1]$.
The formal bilevel problem is
\begin{equation}\label{eqn:bilevel}
\begin{cases}
\psi^*(\theta)=\arg\min_\psi L(\psi,\theta),\\
\theta^*=\arg\min_\theta \frac{1}{M_{\mathrm{meta}}}\sum_{i=1}^{M_{\mathrm{meta}}}\ell_i(\psi^*(\theta)),
\end{cases}
\end{equation}
We solve the problem with a single-loop update~\citep{shu2019meta,yang2024curriculum}.
Writing $g(\theta,\psi)=L(\psi,\theta)$, we replace the exact hypergradient by the first-order penalty surrogate
\begin{equation}\label{eqn:first_order_penalty}
\min_{\theta,\psi}\,
\frac{1}{M_{\mathrm{meta}}}\sum_{i=1}^{M_{\mathrm{meta}}}\ell_i(\psi)
+\lambda^{(t)}\!\left[g(\theta,\psi)-g(\theta,\hat{\psi}^{(t)})\right].
\end{equation}
The first term is the meta-validation loss.
The second encourages approximate first-order optimality of the inner problem.
Since the inner neural objective is nonconvex, Eq.~\eqref{eqn:first_order_penalty} is a surrogate rather than an exact Danskin reduction.
Appendix~\ref{app:first_order_meta_reweighting_convergence} gives an $O(K^{-1/2})$ stationarity term together with an explicit bias floor caused by the inexact inner solution, frozen aggregate, and discrete adjoint.

\noindent\textbf{Weighted Population Dynamics.}
The reweighted state process is
\begin{equation}\label{eqn:mean_field_sde}
\mathrm{d}X_{i,t}=b(X_{i,t},\mu_t,\psi_t)\,\mathrm{d}t+\Sigma\,\mathrm{d}W_{i,t},
\end{equation}
with empirical aggregate
\begin{equation}\label{eqn:weighted_mean_state}
\mu_t=\frac{1}{N}\sum_{i=1}^{N}v_i X_{i,t}.
\end{equation}
We normalize by $1/N$ rather than by $1/\sum_i v_i$, so the absolute scale of the learned weights affects the drift.
For the linear--quadratic specification~\citep{yong2013linear,bensoussan2017linear}, Eq.~\eqref{eqn:mean_field_sde} reduces to
\begin{equation}\label{eqn:lq_drift}
\mathrm{d}X_{i,t}=[a(\mu_t-X_{i,t})+\psi_t] \,\mathrm{d}t+\Sigma\,\mathrm{d}W_{i,t}.
\end{equation}
Euler discretization with Eq.~\eqref{eqn:weighted_mean_state} gives
\begin{align}\label{eqn:discrete_soc}
\min_{\psi,\theta}\frac{1}{N}\!\sum_{i=1}^N\!\!\Big[\!\sum_{s=0}^{S-1}\!R_i(X_{i,s},\mu_s,\psi_s)\Delta t
+\Phi_i(X_{i,S},\mu_S,\psi_S)\!\Big],\nonumber\\
\text{s.t.}\quad
X_{i,s+1}=X_{i,s}+[a(\mu_s-X_{i,s})+\psi_s]\Delta t+\Sigma\Delta W_{i,s},
\quad i\in[N],\,s=0,\ldots,S-1.
\end{align}
The corresponding Hamiltonian is
\begin{equation}\label{eqn:mf_ham}
\mathcal{H}_i(X_{i,t},\mu_t,Y_{i,t},Z_{i,t},\psi_t)
=[a(\mu_t-X_{i,t})+\psi_t]\!\cdot\!Y_{i,t}
+\mathrm{tr}(\Sigma^\top Z_{i,t})
-R_i(X_{i,t},\mu_t,\psi_t).
\end{equation}
Equations~\eqref{eqn:lq_drift}--\eqref{eqn:mf_ham} connect the continuous state model with the discretization used in Algorithm~\ref{alg:pseudo_nddv}.

\noindent\textbf{Relative-State Stability in the Linear--Quadratic Model.}
The shared aggregate and control affect the common motion of the sample states.
Their relative motion has a simpler form because both terms cancel when two state equations are subtracted.
This observation gives an exact stability result rather than an assumed relation between the learned weight range and the final score gap.

Let $G_g\subseteq[N]$, $n_g=|G_g|$, and define
\[
\bar X_{g,t}=\frac1{n_g}\sum_{i\in G_g}X_{i,t},
\qquad
\bar X_t=\frac1N\sum_{i=1}^NX_{i,t}.
\]

\begin{theorem}[Pairwise and group-mean contraction]
\label{thm:relative_state_contraction}
Assume Eq.~\eqref{eqn:lq_drift} with $a>0$, deterministic initial states, and independent standard Wiener processes.
Then, for any $i\ne j$,
\begin{equation}\label{eqn:pairwise_state_contraction}
\mathbb E\|X_{i,t}-X_{j,t}\|^2
=e^{-2at}\|x_i-x_j\|^2
+\frac{1-e^{-2at}}{a}\|\Sigma\|_F^2.
\end{equation}
For any group $G_g$,
\begin{align}
\mathbb E\|\bar X_{g,t}-\bar X_t\|^2
&=e^{-2at}\|\bar x_g-\bar x\|^2 \nonumber\\
&\quad+
\left(\frac1{n_g}-\frac1N\right)
\frac{1-e^{-2at}}{2a}\|\Sigma\|_F^2,
\label{eqn:group_mean_state_contraction}
\end{align}
where $\bar x_g=n_g^{-1}\sum_{i\in G_g}x_i$ and $\bar x=N^{-1}\sum_i x_i$.
The limits at $a=0$ are obtained by continuity.
\end{theorem}

The theorem isolates two effects.
The initial discrepancy decays at rate $a$, while independent diffusion produces a nonzero noise floor.
The weighted mean field $\mu_t$ and the shared control $\psi_t$ do not appear in Eqs.~\eqref{eqn:pairwise_state_contraction}--\eqref{eqn:group_mean_state_contraction} because they are common across sample equations.
The learned weights can still change the common trajectory, the fitted terminal map, and the meta objective, but in this LQ specification they do not directly strengthen relative-state contraction.

\begin{corollary}[Terminal score-gap transfer]
\label{cor:score_gap_transfer}
Condition on the fitted common terminal context $(\mu_T,\psi_T)$ and let
\[
h_i(x)=x\cdot\nabla_x g_i(x)
\]
be the oriented terminal radial score map for data point $i$.
Assume each $h_i$ is $L_h$-Lipschitz with a common deterministic bound and define the map-heterogeneity term
\[
\kappa_{ij}=\sup_x|h_i(x)-h_j(x)|.
\]
Set
\begin{equation}\label{eqn:qij_definition}
Q_{ij}(T)=
\begin{cases}
0, & i=j,\\[2mm]
\left[
 e^{-2aT}\|x_i-x_j\|^2
 +\dfrac{1-e^{-2aT}}{a}\|\Sigma\|_F^2
\right]^{1/2}, & i\neq j.
\end{cases}
\end{equation}
Then the raw group-mean trajectory-score gap satisfies
\begin{equation}\label{eqn:dynamic_group_utility_bound}
\mathbb E|\bar u_g-\bar u|
\le
\frac1{n_gN}\sum_{i\in G_g}\sum_{j=1}^N
\bigl(L_hQ_{ij}(T)+\kappa_{ij}\bigr).
\end{equation}
If the same terminal map is shared across samples, then $\kappa_{ij}=0$.
For the centered NDDV scores,
\begin{equation}\label{eqn:score_disparity}
\mathbb E|\bar\phi_g-\bar\phi|
\le
\frac{N}{N-1}
\frac1{n_gN}\sum_{i\in G_g}\sum_{j=1}^N
\bigl(L_hQ_{ij}(T)+\kappa_{ij}\bigr).
\end{equation}
\end{corollary}

\noindent\textbf{Score Centering.}
Let $\mathcal G=\{G_1,\ldots,G_C\}$ be an evaluation partition.
We use class labels on the main benchmarks and sex in the Adult experiment of Appendix~\ref{app:protected_attribute_validation}.
For any scalar vector $a$, write $\bar a_g=|G_g|^{-1}\sum_{i\in G_g}a_i$ and $\bar a=N^{-1}\sum_i a_i$.

\begin{lemma}[Score-centering identity]
\label{lem:score_centering}
Let $u_i=s_i(\mathcal T)$ and let $\phi_i$ be defined by Eq.~\eqref{eqn:dynamic_dv}.
Then
\begin{equation}\label{eqn:score_centering}
\phi_i=\frac{N}{N-1}(u_i-\bar u),
\qquad
\bar\phi=0.
\end{equation}
Consequently,
\begin{equation}\label{eqn:group_score_identity}
|\bar\phi_g-\bar\phi|
=\frac{N}{N-1}|\bar u_g-\bar u|.
\end{equation}
\end{lemma}

Corollary~\ref{cor:score_gap_transfer} combines this exact affine identity with the state-dynamics bound.
It does not assert that meta-reweighting alone guarantees group balance.
In particular, any empirical reduction in class-conditional score gaps can arise through the common trajectory, the learned terminal map, or the meta objective, whereas the direct relative-state contraction in Eq.~\eqref{eqn:lq_drift} is controlled by $a$ and $\Sigma$.

For the corruption experiments, the positive class is the corruption indicator and the evaluation groups are task labels.
We use the detection true-positive-rate gap (DTPRGap) and detection equalized-odds gap (DEOGap):
\begin{equation}\label{eqn:class_conditional_detection_gaps}
\begin{cases}
\mathrm{DTPRGap}_g=|\mathrm{TPR}-\mathrm{TPR}_g|,\\
\mathrm{DEOGap}_g=|\mathrm{TPR}-\mathrm{TPR}_g|+|\mathrm{FPR}-\mathrm{FPR}_g|.
\end{cases}
\end{equation}
Here TPR and FPR are computed for the fixed detection rule $d_\tau$, and the subscript $g$ restricts that rule to $G_g$.
A group-mean score bound alone does not control a thresholded rate unless the score distribution near $\tau$ is also controlled, so Eq.~\eqref{eqn:class_conditional_detection_gaps} is treated as an empirical diagnostic rather than as a consequence of Corollary~\ref{cor:score_gap_transfer}.

The proofs of Theorem~\ref{thm:relative_state_contraction}, Corollary~\ref{cor:score_gap_transfer}, and Lemma~\ref{lem:score_centering} appear in Appendix~\ref{app:relative_state_stability}.

\begin{figure}[tbp]
    \centering
    \begin{minipage}{0.48\textwidth}
        \centering
        \includegraphics[width=\textwidth,height=0.29\textheight]{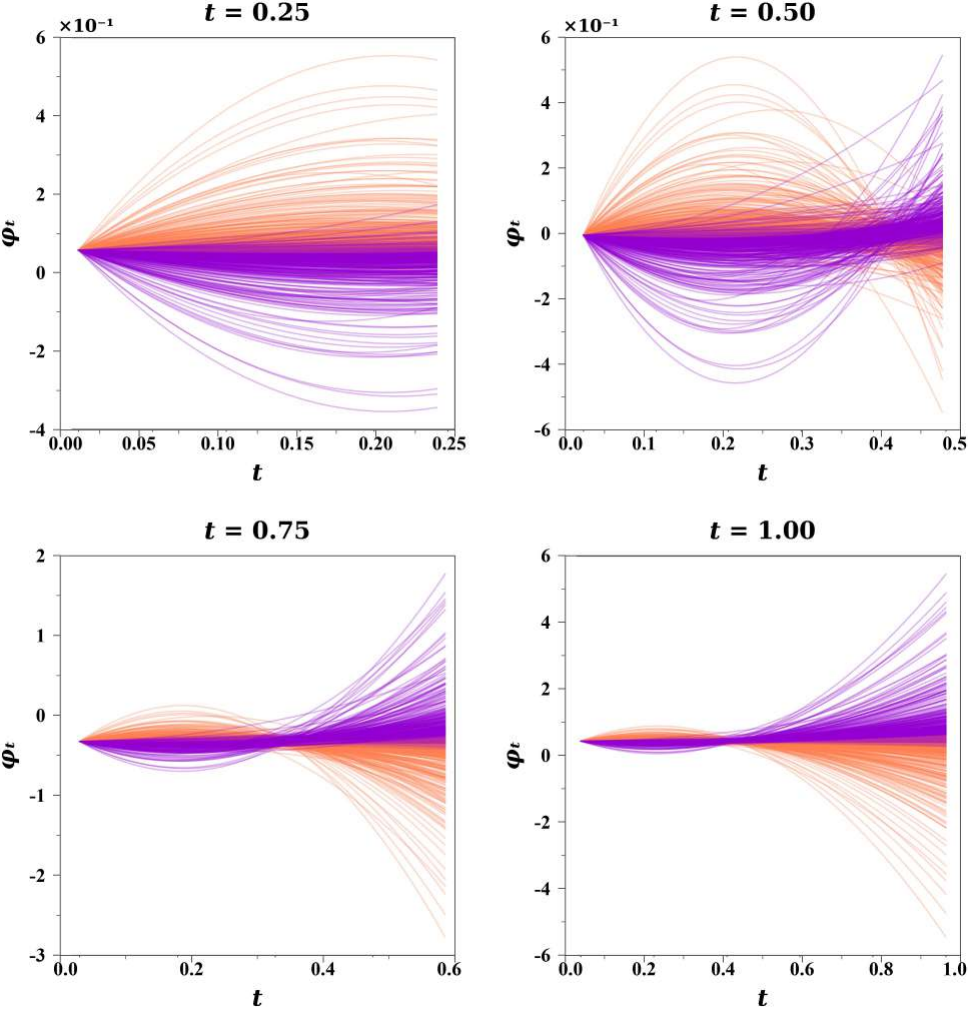}
        \\(a) Layer-wise.
    \end{minipage}%
    \hspace{0.04\textwidth}%
    \begin{minipage}{0.48\textwidth}
        \centering
        \includegraphics[width=\textwidth,height=0.29\textheight]{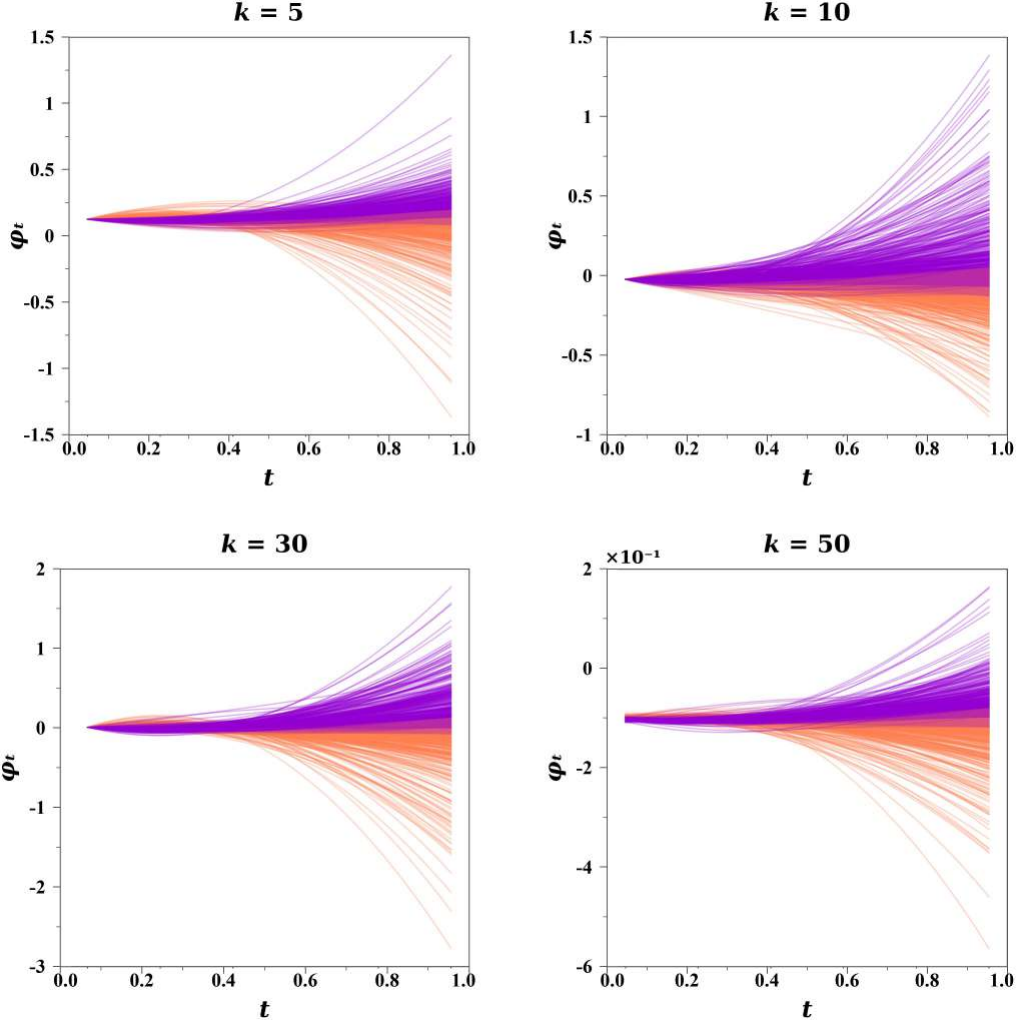}
        \\(b) Epoch-wise.
    \end{minipage}
    \caption{Layer-wise and epoch-wise data value trajectories. a.
    Layer-wise trajectories. b.
    Epoch-wise trajectories.}
    \label{fig:reveal_dv_layer_epoch}
\end{figure}

\begin{figure*}[t]
    \centering
    \includegraphics[width=\textwidth,height=0.54\textheight,keepaspectratio]{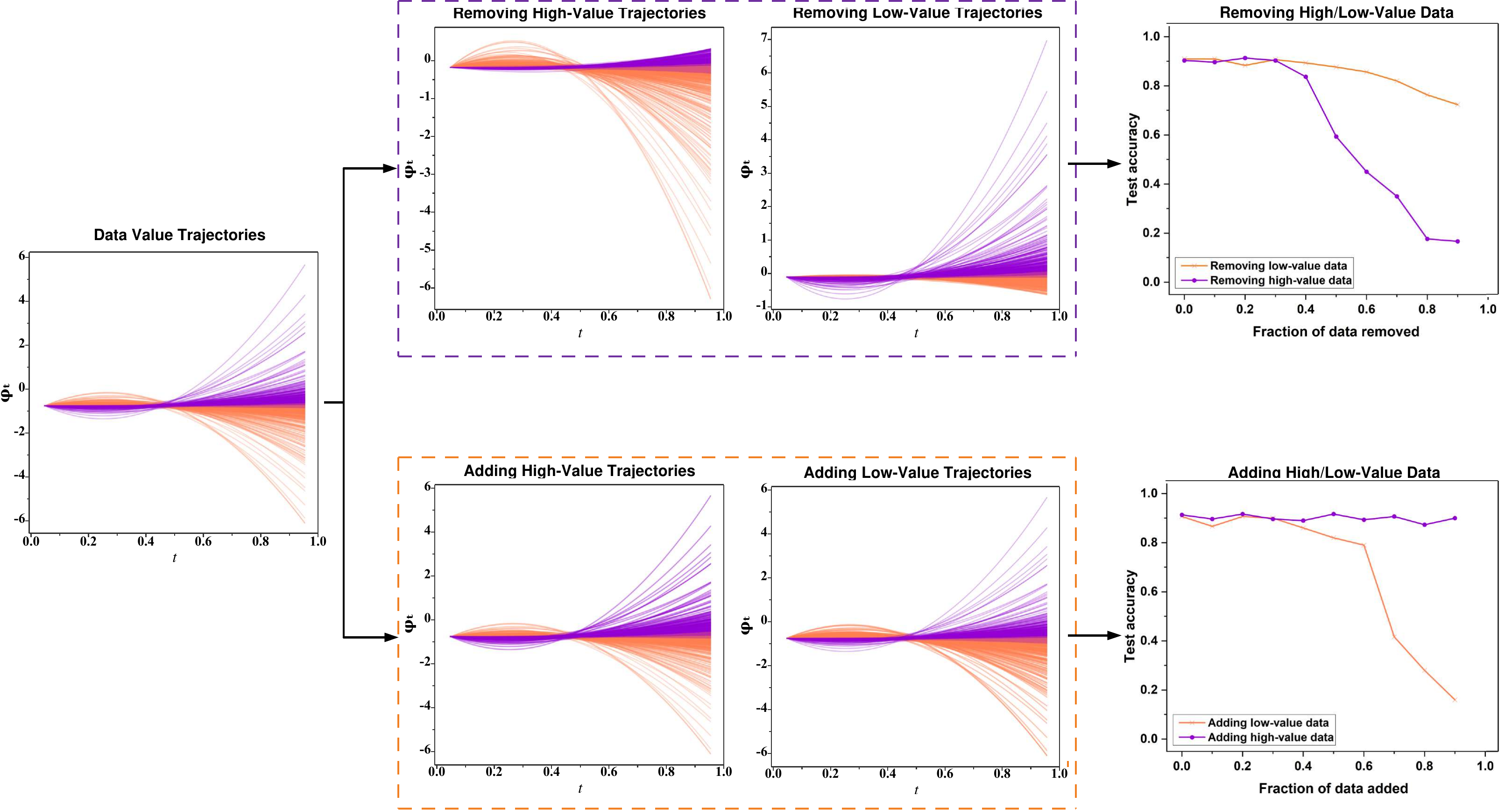}
    \vspace{-1em}
    \caption{Data valuation process on 2dplanes.
    Test accuracy is evaluated after removing or adding high- and low-value trajectories.}
    \label{fig:reveal_dv}
\end{figure*}

\subsection{Structured Parameterization of the Terminal Map}
\label{subsec:interp_nddv}

\begin{figure*}[t]
    \centering
    \includegraphics[width=\textwidth,height=0.72\textheight,keepaspectratio]{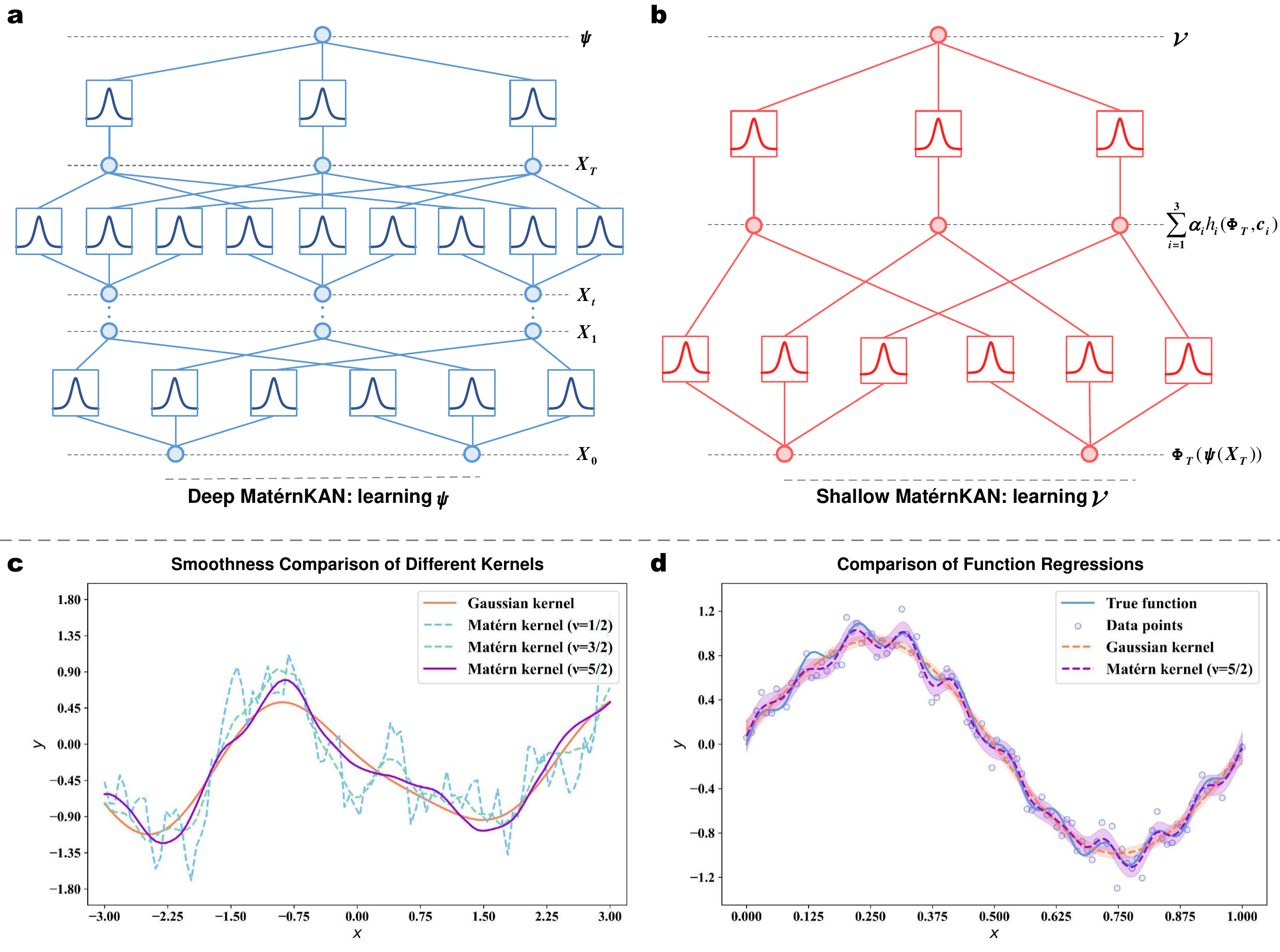}
    \vspace{-1em}
    \caption{Structured terminal-map parameterization with Mat\'ern KAN components.
    Mat\'ernKAN terminal-map structure.}
    \label{fig:terminal_map_interpretability}
\end{figure*}

We parameterize the terminal map with Mat\'ernKANs, which retain the KAN principle of composing learnable univariate functions while replacing generic radial bases with Mat\'ern kernels~\citep{liu2024kan}.
In this section, a deep Mat\'ernKAN represents the trajectory control $\psi$, and a shallow Mat\'ernKAN represents the terminal valuator $\mathcal V$.
See Figure~\ref{fig:terminal_map_interpretability}(a,b).
The kernel-smoothness and regression comparisons in Figure~\ref{fig:terminal_map_interpretability}(c,d) motivate this choice and support the use of inspectable coordinate-wise terminal-map components.

For $X_t=(X_{t,1},\ldots,X_{t,d})$, the scalar function $f_\vartheta$ is expressed as a composition of learned univariate maps:
\begin{equation}\label{eqn:kan_control}
f_{\vartheta}(X_t)=\sum_{q=1}^{2d+1} H_q\!\left(\sum_{p=1}^{d}h_{q,p}(X_{t,p})\right),
\end{equation}
where $h_{q,p}:\mathbb R\to\mathbb R$ and $H_q:\mathbb R\to\mathbb R$ are learned univariate functions.
We parameterize each inner function as
\begin{equation}\label{eqn:rbf}
h_{q,p}(X_{t,p})=\alpha_b h_b(X_{t,p})+\alpha_k h_k(X_{t,p},c),
\end{equation}
where $h_b$ is SiLU, $h_k$ is a radial basis centered at $c$, and $\alpha_b,\alpha_k$ use Xavier initialization.
The radial term is a Mat\'ern kernel~\citep{rasmussen2003gaussian}:
\begin{equation}\label{eqn:matern_kernel}
h_k(x,c)
=\frac{2^{1-\nu}}{\Gamma(\nu)}
\left(\frac{\sqrt{2\nu}\,|x-c|}{\ell}\right)^{\!\nu}
K_\nu\!\left(\frac{\sqrt{2\nu}\,|x-c|}{\ell}\right),
\end{equation}
where $\ell$ is the length scale, $\nu>0$ controls smoothness, $\Gamma$ is the Gamma function, and $K_\nu$ is the modified Bessel function.
The Gaussian limit is recovered as $\nu\to\infty$.
At $\nu=3/2$,
\begin{equation}\label{eqn:matern_three_halves}
h_k^{3/2}(x,c)
=\left(1+\frac{\sqrt{3}|x-c|}{\ell}\right)
\exp\!\left(-\frac{\sqrt{3}|x-c|}{\ell}\right).
\end{equation}
This decomposition makes the coordinate-wise functions of the fitted terminal map directly inspectable.
We interpret them as features of the learned representation, not as causal effects in the absence of intervention.

\section{Computational Cost of One-Run Valuation}\label{sec:complexity}

The computational claim concerns the cost of releasing trajectory-conditioned marginal-contribution scores from one fitted forward--backward system.
We separate model fitting, score assignment, and the optional retraining diagnostics used to compare NDDV with coalition values.
Let $n$ be the number of training data points, $B$ the mini-batch size, $E$ the number of passes through the training set, and $S$ the number of Euler steps; let $W$, $G$, and $d$ denote the hidden width, the number of KAN basis functions, and the representation dimension.
Constants associated with the meta-network and automatic differentiation are absorbed below.

\begin{itemize}
    \item \textbf{Fitting the trajectory model.} A mini-batch update propagates $B$ states through $S$ forward and backward steps.
    Under the parameterization in Section~\ref{subsec:interp_nddv}, its arithmetic cost is
    \[
        C_{\mathrm{batch}}=O\!\left(BSW^2G+BdW\right).
    \]
    One epoch contains $\lceil n/B\rceil$ updates, so $E$ epochs cost
    \begin{equation}\label{eqn:nddv_training_complexity}
        O\!\left(En(SW^2G+dW)\right).
    \end{equation}
    For fixed architecture, integration horizon, and epoch budget, this cost is linear in $n$.

    \item \textbf{Assigning NDDV scores.} After fitting, a batched forward--adjoint pass evaluates the terminal state and adjoint for each data point and applies Eq.~\eqref{eqn:dynamic_dv}.
    The cost is $O(n(SW^2G+dW))$ and no predictor is retrained for an individual data point or coalition.
    Averaging $M_{\mathrm{traj}}$ independent stochastic paths multiplies this scoring cost by $M_{\mathrm{traj}}$.
    The wall-clock comparison uses the single-path computation in Algorithm~\ref{alg:pseudo_nddv}.

    \item \textbf{Optional ordering diagnostics.} Theorem~\ref{thm:ranking_consistency} does not change the cost of computing NDDV.
    Estimating its pair-specific error is a separate validation procedure.
    With $P$ pre-specified pairs and $M_{\mathrm{coal}}$ sampled coalitions per pair, the dominant additional cost is $O(PM_{\mathrm{coal}}C_{\mathrm{pred}})$, where $C_{\mathrm{pred}}$ is the cost of fitting the downstream predictor on one coalition.
    Coalition fits shared across pairs can be cached.
    We exclude this diagnostic cost from the NDDV runtime and give its sampling protocol separately.
\end{itemize}

The corresponding costs of comparison methods depend on their evaluation budgets.
KNNShapley computes distances and sorts the training data points for each validation query.
Coalition methods scale with the number of sampled subsets times the predictor-fitting cost.
Data-OOB scales with the number and size of its fitted trees.
Because these constants and budgets differ substantially, Section~\ref{subsec:runtime_efficiency} emphasizes measured wall-clock time under fixed configurations.
Linear sample-size scaling of Eq.~\eqref{eqn:nddv_training_complexity} is a computational statement only.
Agreement with a static value is governed by the approximation terms in Theorem~\ref{thm:ranking_consistency}.

\section{Experiments}\label{sec:experiments}
The experiments evaluate three aspects of NDDV: fidelity to finite retraining marginals when such marginals are estimable, computational cost for releasing valuation scores, and utility of the induced rankings for corrupted-sample detection and data curation.
Class-conditional, noise-level, and sensitivity diagnostics then characterize when the fitted trajectory model is stable and when the results require more cautious interpretation.

\subsection{Experimental Setup}\label{subsec:exper_setup}
Table~\ref{tab:summary_of_datasets} lists the tabular, text, and image benchmarks~\citep{jiang2023opendataval,deng2009imagenet}.
Methods use the same fixed representations and downstream logistic-regression predictor, so the comparison isolates the valuation rule.

The baselines cover coalition-based, influence-based, ensemble, geometry-based, and run-specific valuation.
They include LOO~\citep{koh2017understanding}, Data Shapley~\citep{ghorbani2019}, Beta Shapley~\citep{kwon2022beta}, Data Banzhaf~\citep{wang2022data}, InfluenceFunction~\citep{feldman2020neural}, DVRL~\citep{yoon2020data}, KNNShapley~\citep{jia2019b}, AME~\citep{lin2022measuring}, Data-OOB~\citep{kwon2023dataoob}, DU-Shapley~\citep{garrido2024shapley}, LAVA~\citep{just2023lava}, and GhostSuite~\citep{wang2024data}.
Corrupted-sample detection uses a budget-matched threshold, so each method flags the same fraction of data points.
Unless stated otherwise, means and standard deviations are computed over five independently reseeded runs.

For retraining-based comparisons, the reference utility is validation accuracy of the downstream predictor trained on the selected coalition.
Because the theory is stated for smooth validation surrogates, these accuracy-based marginals serve as an empirical stress test of the ranking behavior.
Class-conditional diagnostics use task labels as groups and are not demographic fairness claims.
Implementation details are provided in the released code at \url{https://github.com/liangzhangyong/NDDV}.

\begin{table}
    \centering
\caption{Datasets used in the experiments.}
    \label{tab:summary_of_datasets}
    \resizebox{0.95\textwidth}{!}{
    \begin{tabular}{c|cccccc}
        \toprule
        \multirow{2}{*}{\shortstack{Dataset}}
        & Sample & Input & Number of & Minor Class & Data & Group \\
        & Size & Dimension & Classes & Proportion & Type & Type \\
        \midrule
        2dplanes \citep{feurer-arxiv19a} & 40768 & 10 & 2 & 0.499 & Tabular & Class label \\
        electricity \citep{gama2004learning} & 38474 & 6 & 2 & 0.5 & Tabular & Class label \\
        BBC \citep{greene2006practical} & 2225 & 768 & 5 & 0.17 & Text & Class label \\
        IMDB \citep{maas2011learning} & 50000 & 768 & 2 & 0.5 & Text & Class label \\
        STL10 \citep{coates2011analysis} & 5000 & 2048 & 10 & 0.10 & Image & Class label \\
        CIFAR10 \citep{krizhevsky2009learning} & 50000 & 2048 & 10 & 0.1 & Image & Class label \\
        ImageNet100 \citep{deng2009imagenet} & 130000 & 768 & 100 & 0.01 & Image & Class label \\
        \bottomrule
    \end{tabular}
    }
\end{table}

\subsection{Fidelity and Efficiency}
\label{subsec:empirical_epsilon}
\label{subsec:runtime_efficiency}

\vspace{0.5em}
\noindent\textbf{Finite-marginal fidelity.}
We first test whether the trajectory score preserves the ordering induced by finite retraining marginals.
Since exhaustive coalition retraining is infeasible at benchmark scale, we use a conservative coalition audit on three datasets.
Table~\ref{tab:epsilon_delta} gives the reference-coalition discrepancy $\hat\varepsilon_{\mathrm{ref}}$ and the context-variation term $\hat\delta_{\mathrm{ctx}}$.
Their combination gives the empirical pair-specific error term used to interpret Theorem~\ref{thm:ranking_consistency}.

\begin{table}[tbp]
\centering
\begingroup
\normalsize
\caption{Empirical finite-marginal error terms.}
\label{tab:epsilon_delta}
\begin{tabular}{ccc}
\toprule
Dataset & $\hat\varepsilon_{\mathrm{ref}}$ & $\hat\delta_{\mathrm{ctx}}$ \\
\midrule
2dplanes & $0.032 \pm 0.011$ & $0.048 \pm 0.019$ \\
BBC & $0.041 \pm 0.015$ & $0.063 \pm 0.024$ \\
CIFAR10 & $0.057 \pm 0.021$ & $0.089 \pm 0.031$ \\
\bottomrule
\end{tabular}
\endgroup
\end{table}

\begin{table}[tbp]
\centering
\caption{Normalized pairwise LOO value-gap distributions.}\label{tab:value_gap_distribution}
\resizebox{0.75\textwidth}{!}{
\begin{tabular}{ccccc}
\toprule
Dataset & Median gap & 75th pct. & 90th pct. & $2\hat\varepsilon_{\mathrm{ref}}+\hat\delta_{\mathrm{ctx}}$ \\
\midrule
2dplanes & $0.18$ & $0.31$ & $0.52$ & $0.112$ \\
BBC      & $0.21$ & $0.36$ & $0.61$ & $0.145$ \\
CIFAR10  & $0.26$ & $0.44$ & $0.73$ & $0.203$ \\
\midrule
2dplanes (clean) & $0.08$ & $0.15$ & $0.28$ & $0.112$ \\
BBC (clean)      & $0.09$ & $0.17$ & $0.31$ & $0.145$ \\
CIFAR10 (clean)  & $0.11$ & $0.21$ & $0.38$ & $0.203$ \\
\bottomrule
\end{tabular}}
\end{table}

The resulting envelopes range from $0.112$ to $0.203$.
In the noisy regime, the $90$th-percentile LOO gap in Table~\ref{tab:value_gap_distribution} exceeds the envelope across the evaluated datasets.
The clean gaps are much smaller.
This matches the theory: ordering agreement is expected when finite marginal gaps are large relative to local approximation error.

Table~\ref{tab:rank_corr} gives the same pattern from a rank-correlation viewpoint.
NDDV correlates strongly with exact LOO and sampled Shapley rankings under $10\%$ label noise, while the correlation weakens or disappears on clean data.
The clean case reflects a regime where the sufficient ordering condition is largely inactive.

\begin{table}[tbp]
\centering
\caption{Spearman correlations with retraining-based rankings.}
\label{tab:rank_corr}
\begin{tabular}{ccc}
\toprule
Dataset & $\rho$(NDDV, LOO) & $\rho$(NDDV, Shapley) \\
\midrule
\multicolumn{3}{l}{\emph{Noisy regime ($10\%$ label noise)}} \\
2dplanes & $0.91 \pm 0.02$ & $0.89 \pm 0.03$ \\
BBC & $0.87 \pm 0.03$ & $0.85 \pm 0.04$ \\
CIFAR10 & $0.82 \pm 0.04$ & $0.79 \pm 0.05$ \\
\midrule
\multicolumn{3}{l}{\emph{Clean regime (no injected noise)}} \\
2dplanes & $-0.04 \pm 0.05$ & $-0.02 \pm 0.06$ \\
BBC & $\phantom{-}0.31 \pm 0.06$ & $\phantom{-}0.28 \pm 0.07$ \\
CIFAR10 & $\phantom{-}0.46 \pm 0.05$ & $\phantom{-}0.42 \pm 0.06$ \\
\bottomrule
\end{tabular}
\end{table}

The fidelity experiments show that NDDV tracks retraining-based orderings when harmful samples create clear marginal gaps.
On benign data, the smaller marginal gaps make the ranking agreement weaker.

\begin{figure*}[t]
    \centering
    \includegraphics[width=\textwidth,height=0.72\textheight,keepaspectratio]{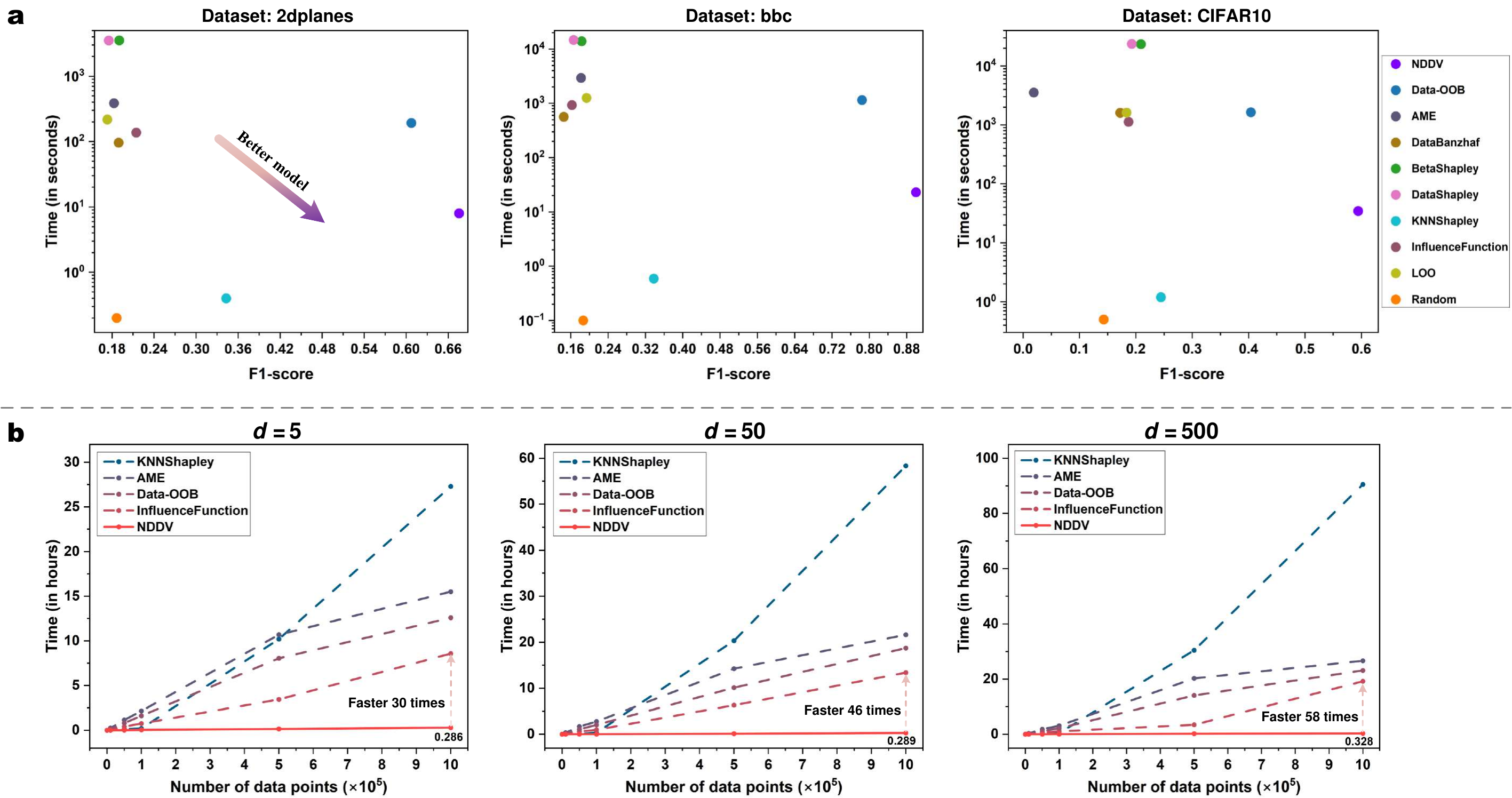}
    \vspace{-1em}
    \caption{Elapsed time for releasing valuation scores under the fixed benchmarks.}
    \label{fig:time_comparison}
\end{figure*}

\vspace{0.5em}
\noindent\textbf{Cost of releasing valuation scores.}
Figure~\ref{fig:time_comparison} and Table~\ref{tab:wallclock_scaling} measure the cost of producing valuation scores under the same benchmark pipeline.
On the fixed configurations, NDDV is among the fastest methods because it avoids repeated coalition retraining.
On synthetic scaling tests, its runtime grows near-linearly with $n$ under the fixed architecture.
At the largest evaluated setting, $(n,d)=(10^6,500)$, NDDV takes $18$ minutes, whereas KNN-Shapley takes $1{,}060\pm30$ minutes and Data Shapley times out.
This $58\times$ ratio reflects the stated architecture, budget, and hardware.

\begin{table}[tbp]
\centering
    \caption{Wall-clock scaling on synthetic data ($d=500$).}\label{tab:wallclock_scaling}
\begin{tabular}{ccccc}
\toprule
$n$ & NDDV (min) & KNN-Shapley (min) & Speed-up & Data Shapley \\
\midrule
$10^4$ & $1.6$ & $6.1$    & $\sim 4\times$  & $\sim 24$\,h \\
$10^5$ & $5.2$ & $58$     & $\sim 11\times$ & $>24$\,h (timeout) \\
$10^6$ & $18$  & $1{,}060\pm 30$ & $\mathbf{58\times}$ & $>24$\,h (timeout) \\
\bottomrule
\end{tabular}
\end{table}


\subsection{Corruption Detection and Data-Curation Utility}
\label{subsec:f1_score_comparison}

\vspace{0.5em}
\noindent\textbf{Corrupted-sample detection.}
Tables~\ref{tab:mislabeled_data_detection} and~\ref{tab:mislabeled_data_detection_n10000} give F1 scores for detecting injected corruption.
At $n=1{,}000$, NDDV has the top F1-score on six of seven rows, with Data-OOB higher on electricity by $0.01$.
At $n=10{,}000$, NDDV gives the top F1-score in each evaluated row.
The improvement varies across datasets, with smaller margins on electricity and CIFAR10 and larger margins on BBC and STL10, indicating sensitivity to representation quality and corruption structure.

Data-OOB is the strongest non-NDDV baseline at $n=1{,}000$, and GhostSuite is the closest trajectory-aware comparator on several datasets.
AME performs poorly under the fixed-budget threshold.
F1 measures corruption-detection utility for the resulting rankings.

\begin{table*}[t]
    \centering
    \caption{Corrupted-sample detection F1-score at $n=1{,}000$.}
    \label{tab:mislabeled_data_detection}
    \resizebox{\textwidth}{!}{%
    \begin{tabular}{c|cccccccccccc}
        \toprule
        \multirow{2}{*}{Dataset} & \multirow{2}{*}{LOO} & Data & Beta & Data & Influence & KNN & \multirow{2}{*}{AME} & Data & \multirow{2}{*}{LAVA} & DU- & \multirow{2}{*}{GhostSuite} & \multirow{2}{*}{NDDV} \\
        & & Shapley & Shapley & Banzhaf & Function & Shapley & & -OOB & & Shapley & & \\
        \midrule
        \multirow{2}{*}{2dplanes} & $0.18 \pm $ & $0.17 \pm$ & $0.16 \pm$ & $0.16 \pm$ & $0.18 \pm$ & $0.30 \pm$ & $0.18 \pm$ & $0.46 \pm$ & $0.21 \pm$ & $0.33 \pm$ & $\underline{0.52} \pm$ & $\mathbf{0.67} \pm$ \\
        & $0.003$ & $0.005$ & $0.003$ & $0.009$ & $0.005$ & $0.007$ & $0.009$ & $0.007$ & $0.008$ & $0.006$ & $0.006$ & $0.005$ \\
        \addlinespace[1mm]
        \multirow{2}{*}{electricity} & $0.18 \pm$ & $0.17 \pm$ & $0.19 \pm$ & $0.18 \pm$ & $0.19 \pm$ & $0.23 \pm$ & $0.01 \pm$ & $\mathbf{0.37} \pm$ & $0.20 \pm$ & $0.26 \pm$ & $0.34 \pm$ & $\underline{0.36} \pm$ \\
        & $0.004$ & $0.004$ & $0.006$ & $0.002$ & $0.003$ & $0.006$ & $0.010$ & $0.002$ & $0.007$ & $0.005$ & $0.003$ & $0.002$ \\
        \addlinespace[1mm]
        \multirow{2}{*}{BBC} & $0.12 \pm$ & $0.11 \pm$ & $0.11 \pm$ & $0.18 \pm$ & $0.16 \pm$ & $0.31 \pm$ & $0.11 \pm$ & $0.18 \pm$ & $0.24 \pm$ & $0.35 \pm$ & $\underline{0.49} \pm$ & $\mathbf{0.86} \pm$ \\
        & $0.004$ & $0.004$ & $0.003$ & $0.005$ & $0.002$ & $0.008$ & $0.009$ & $0.004$ & $0.009$ & $0.006$ & $0.005$ & $0.002$ \\
        \addlinespace[1mm]
        \multirow{2}{*}{IMDB} & $0.12 \pm$ & $0.09 \pm$ & $0.09 \pm$ & $0.15 \pm$ & $0.16 \pm$ & $0.22 \pm$ & $0.18 \pm$ & $0.17 \pm$ & $0.19 \pm$ & $0.21 \pm$ & $\underline{0.24} \pm$ & $\mathbf{0.27} \pm $ \\
        & $0.002$ & $0.004$ & $0.003$ & $0.002$ & $0.009$ & $0.008$ & $0.011$ & $0.005$ & $0.008$ & $0.006$ & $0.005$ & $0.007$ \\
        \addlinespace[1mm]
        \multirow{2}{*}{STL10} & $0.13 \pm$ & $0.17 \pm$ & $0.16 \pm$ & $0.18 \pm$ & $0.14 \pm$ & $0.28 \pm$ & $0.01 \pm$ & $0.22 \pm$ & $0.25 \pm$ & $0.30 \pm$ & $\underline{0.41} \pm$ & $\mathbf{0.71} \pm$ \\
        & $0.006$ & $0.004$ & $0.002$ & $0.005$ & $0.009$ & $0.007$ & $0.009$ & $0.003$ & $0.008$ & $0.005$ & $0.007$ & $0.008$ \\
        \addlinespace[1mm]
        \multirow{2}{*}{CIFAR10} & $0.18 \pm$ & $0.19 \pm$ & $0.20 \pm$ & $0.17 \pm$ & $0.19 \pm$ & $0.24 \pm$ & $0.02 \pm$ & $0.40 \pm$ & $0.22 \pm$ & $0.29 \pm$ & $\underline{0.47} \pm$ & $\mathbf{0.59} \pm$ \\
        & $0.004$ & $0.003$ & $0.005$ & $0.002$ & $0.007$ & $0.004$ & $0.008$ & $0.004$ & $0.009$ & $0.005$ & $0.005$ & $0.004$ \\
        \addlinespace[1mm]
        \multirow{2}{*}{ImageNet100}
        & $0.08 \pm$ & $0.87 \pm$ & $0.87 \pm$ & $0.08 \pm$ & $0.08 \pm$ & $0.67 \pm$
        & $0.11 \pm$ & $\underline{0.95} \pm$ & $0.08 \pm$ & $0.06 \pm$ & $0.34 \pm$
        & $\mathbf{0.99} \pm$ \\
        & $0.004$ & $0.004$ & $0.004$ & $0.004$ & $0.006$ & $0.007$ & $0.009$ & $0.004$ & $0.008$ & $0.006$ & $0.005$ & $0.005$ \\
        \bottomrule
    \end{tabular}
    }
\end{table*}

\begin{table*}[t]
    \centering
    \caption{Corrupted-sample detection F1-score at $n=10{,}000$.}
    \label{tab:mislabeled_data_detection_n10000}
    \resizebox{\textwidth}{!}{%
    \begin{tabular}{c|ccccccc}
        \toprule
        \multirow{2}{*}{Dataset} & KNN & \multirow{2}{*}{AME} & Data & \multirow{2}{*}{LAVA} & DU- & \multirow{2}{*}{GhostSuite} & \multirow{2}{*}{NDDV} \\
        & Shapley & & -OOB & & Shapley & & \\
        \midrule
        2dplanes & $0.37 \pm 0.004$ & $0.01 \pm 0.012$ & $0.71 \pm 0.002$ & $0.28 \pm 0.007$ & $0.42 \pm 0.005$ & $\underline{0.64} \pm 0.004$ & $\mathbf{0.79} \pm 0.005$ \\
        \addlinespace[1mm]
        electricity & $0.32 \pm 0.001$ & $0.01 \pm 0.009$ & $0.38 \pm 0.003$ & $0.25 \pm 0.006$ & $0.35 \pm 0.004$ & $\underline{0.41} \pm 0.003$ & $\mathbf{0.44} \pm 0.002$ \\
        \addlinespace[1mm]
        BBC & $0.52 \pm 0.005$ & $0.01 \pm 0.010$ & $0.73 \pm 0.002$ & $0.31 \pm 0.008$ & $0.48 \pm 0.005$ & $\underline{0.62} \pm 0.004$ & $\mathbf{0.85} \pm 0.006$ \\
        \addlinespace[1mm]
        IMDB & $0.29 \pm 0.002$ & $0.18 \pm 0.012$ & $\underline{0.48} \pm 0.002$ & $0.23 \pm 0.007$ & $0.33 \pm 0.004$ & $0.45 \pm 0.003$ & $\mathbf{0.52} \pm 0.003$ \\
        \addlinespace[1mm]
        STL10 & $0.16 \pm 0.009$ & $0.01 \pm 0.012$ & $0.77 \pm 0.002$ & $0.32 \pm 0.007$ & $0.45 \pm 0.004$ & $\underline{0.59} \pm 0.005$ & $\mathbf{0.91} \pm 0.003$ \\
        \addlinespace[1mm]
        CIFAR10 & $0.27 \pm 0.009$ & $0.01 \pm 0.010$ & $0.46 \pm 0.001$ & $0.26 \pm 0.008$ & $0.37 \pm 0.004$ & $\underline{0.51} \pm 0.003$ & $\mathbf{0.58} \pm 0.004$ \\
        \addlinespace[1mm]
        ImageNet100 & $0.89 \pm 0.005$ & $0.11 \pm 0.011$ & $\underline{0.94} \pm 0.002$ & $0.10 \pm 0.007$ & $0.10 \pm 0.004$ & $0.43 \pm 0.004$ & $\mathbf{0.98} \pm 0.004$ \\
        \bottomrule
    \end{tabular}
    }
\end{table*}

The semivalue baselines in Table~\ref{tab:mislabeled_data_detection} provide the matched effectiveness comparison used in this section.
The ImageNet100 rows summarize the larger-scale settings under the same fixed-budget detector.

\begin{table*}[t]
\centering
\caption{ImageNet100 scaling results.}
\label{tab:imagenet100_completed_diagnostics}
\resizebox{\textwidth}{!}{%
\begin{tabular}{cccccc}
\toprule
Evaluation & Setting & Metric & Best baseline & NDDV & $\Delta$ \\
\midrule
Head-to-head & $10\%$ label noise, $n=7{,}500$ & F1 & Data-OOB: $0.979$ & $\mathbf{0.980}$ & $+0.001$ \\
Label-noise mean & Noise rates $5\%$--$45\%$ & F1 & Data-OOB: $\mathbf{0.937}$ & $0.933$ & $-0.004$ \\
Feature-noise mean & Noise rates $5\%$--$45\%$ & F1 & Data-OOB: $0.806$ & $\mathbf{0.864}$ & $+0.058$ \\
\midrule
Scale & $n=100$ & F1 & GhostSuite: $1.000$ & $1.000$ & $0.000$ \\
Scale & $n=100$ & Recall-AUC & GhostSuite: $0.950$ & $0.950$ & $0.000$ \\
Scale & $n=1{,}000$ & F1 & GhostSuite: $0.452$ & $\mathbf{0.985}$ & $+0.533$ \\
Scale & $n=1{,}000$ & Recall-AUC & GhostSuite: $0.950$ & $0.950$ & $0.000$ \\
Scale & $n=10{,}000$ & F1 & LossVal: $0.860$ & $\mathbf{0.980}$ & $+0.120$ \\
Scale & $n=10{,}000$ & Recall-AUC & GhostSuite: $0.950$ & $0.950$ & $0.000$ \\
\bottomrule
\end{tabular}
}
\end{table*}

\vspace{0.5em}
\noindent\textbf{Value-directed data selection.}
\label{subsec:robustness_corruption_manipulation}

\begin{figure*}[t]
    \centering
    \includegraphics[width=\textwidth,height=0.72\textheight,keepaspectratio]{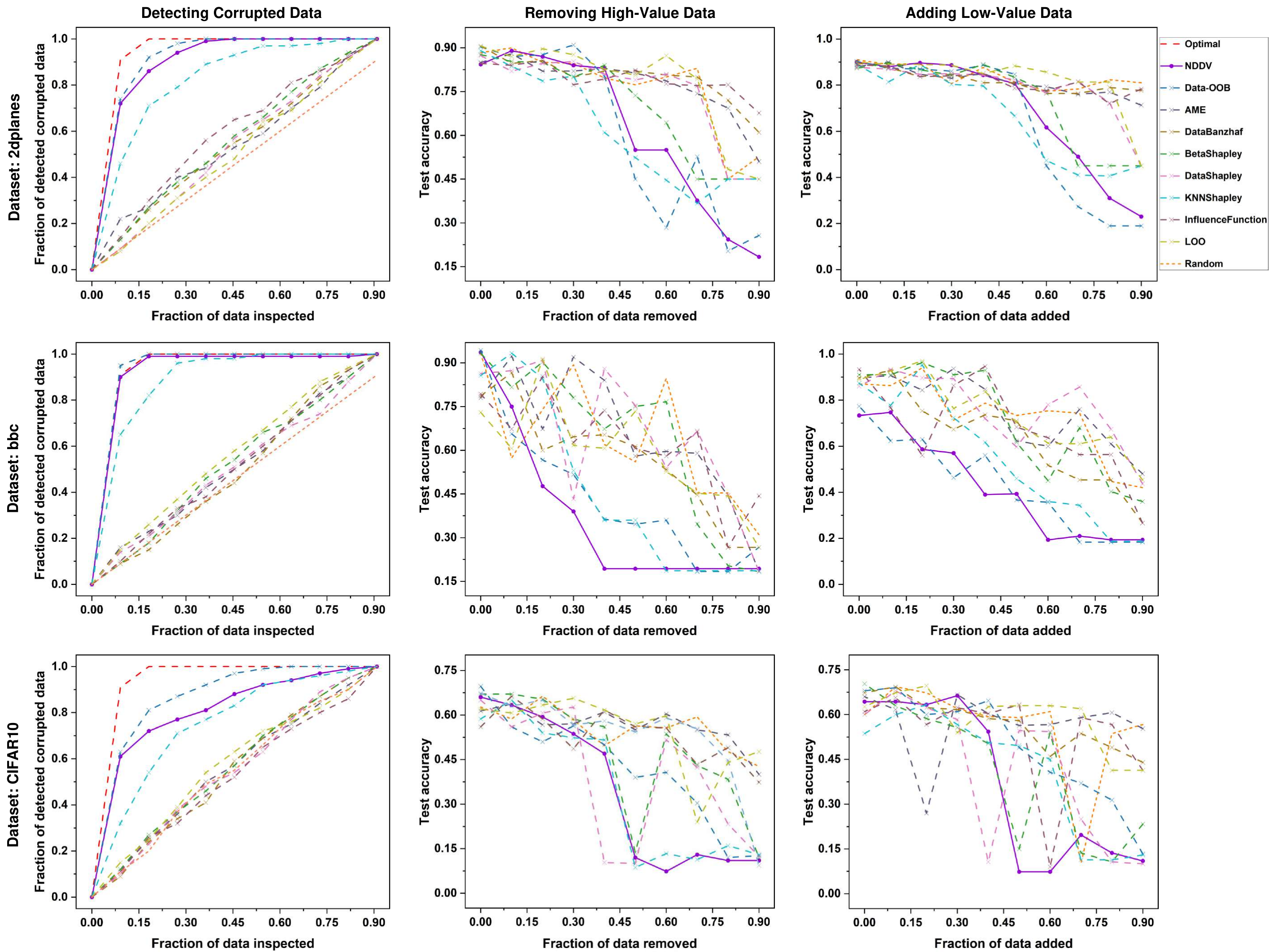}
    \vspace{-1em}
    \caption{Value-directed data manipulation under 10\% label noise.
    Columns show detection, removal, and addition tasks.}
    \label{fig:corrupt_remove_add_exp}
\end{figure*}

Figure~\ref{fig:corrupt_remove_add_exp} evaluates whether the NDDV ranking supports value-directed data selection under $10\%$ label corruption.
Removing high-valued data points should reduce test accuracy quickly, whereas adding low-valued data points first should slow performance recovery.
The NDDV curves show sharp accuracy drops when high-valued data points are removed and slower recovery when low-valued data points are added first.
These experiments evaluate the utility of the induced ranking for data curation.

\begin{figure*}[t]
    \centering
    \includegraphics[width=\textwidth,height=0.72\textheight,keepaspectratio]{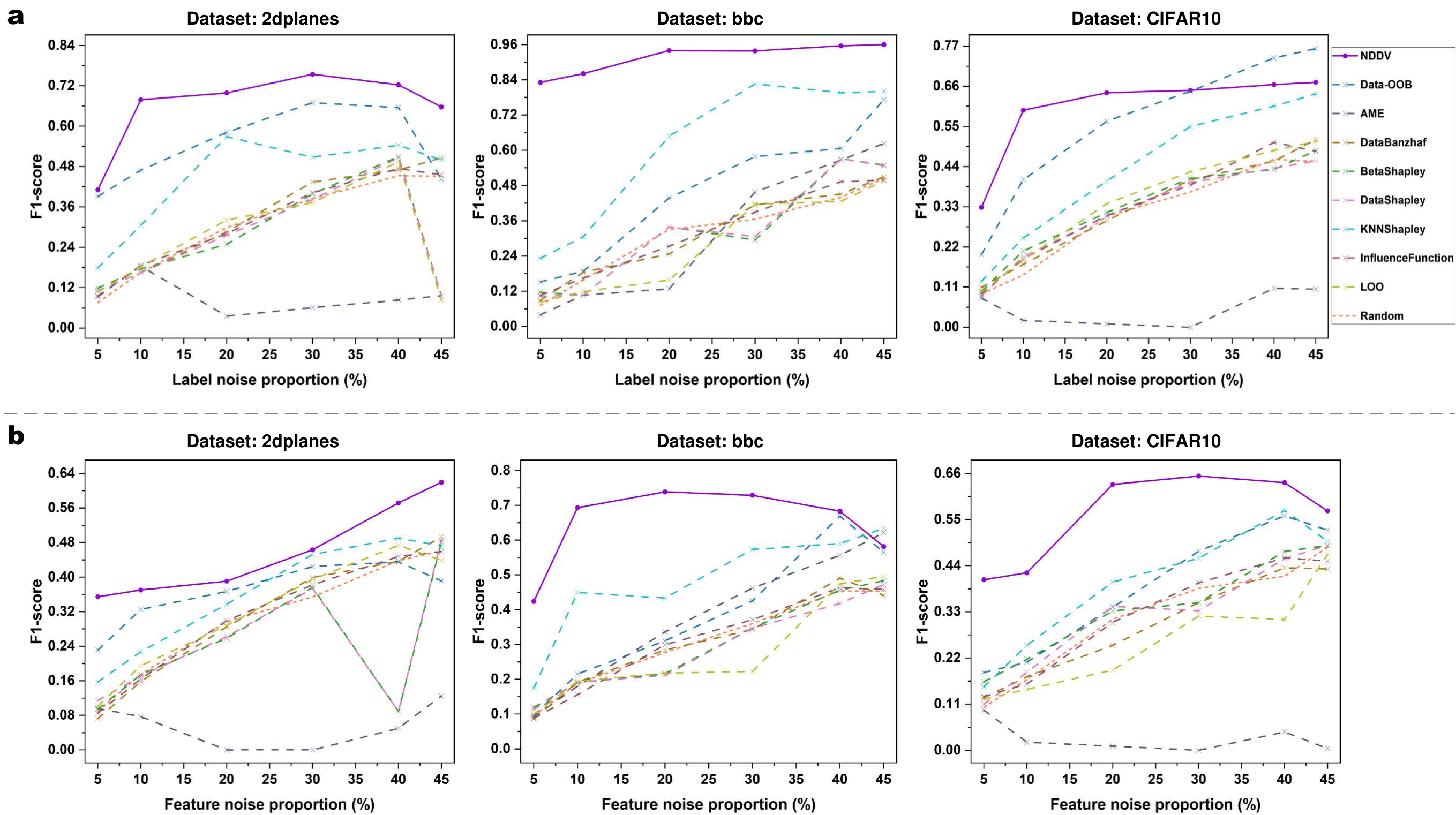}
    \vspace{-1em}
    \caption{Noisy data detection on six benchmarks. a.
    Noisy-label detection. b.
    Noisy-feature detection.
    F1-score across six noise levels.}
    \label{fig:feature_label_noise_exp}
\end{figure*}

\vspace{0.5em}
\noindent\textbf{Noise robustness.}
Tables~\ref{tab:label_noise_proportion} and~\ref{tab:feature_noise_proportion}, together with Fig.~\ref{fig:feature_label_noise_exp}, vary the corruption rate from $5\%$ to $45\%$.
Under label noise, NDDV has the highest F1-score at $5\%$, $10\%$, $40\%$, and $45\%$, while Data-OOB is higher at $20\%$ and $30\%$.
Under feature noise, NDDV has the highest F1-score at each evaluated noise rate.
The relative ordering depends on both corruption type and severity.

\begin{table}[tbp]
    \centering
    \caption{F1-score under different label-noise rates.}
    \label{tab:label_noise_proportion}
    \resizebox{\textwidth}{!}{%
    \begin{tabular}{c|ccccccccc} 
        \toprule 
        \multirow{2}{*}{\shortstack{Noise\\Rate}} 
         & \multirow{2}{*}{LOO} & Data & Beta & Data & Influence & KNN & \multirow{2}{*}{AME} & Data & \multirow{2}{*}{NDDV} \\ 
         & & Shapley & Shapley & Banzhaf & Function & Shapley & & -OOB & \\
         \midrule 
         5\% & $0.09 \pm 0.003$ & $0.12 \pm 0.007$ & $0.11 \pm 0.008$ & $0.09 \pm 0.004$ & $0.11 \pm 0.003$ & $0.17 \pm 0.003$ & $0.01 \pm 0.009$ & $0.62 \pm 0.002$ & $\mathbf{0.74} \pm 0.003$ \\
         10\% & $0.16 \pm 0.007$ & $0.19 \pm 0.010$ & $0.19 \pm 0.009$ & $0.18 \pm 0.005$ & $0.18 \pm 0.003$ & $0.30 \pm 0.003$ & $0.18 \pm 0.010$ & $0.74 \pm 0.002$ & $\mathbf{0.76} \pm 0.003$ \\
         20\% & $0.30 \pm 0.005$ & $0.25 \pm 0.008$ & $0.25 \pm 0.008$ & $0.31 \pm 0.002$ & $0.31 \pm 0.002$ & $0.45 \pm 0.004$ & $0.010 \pm 0.009$ & $\mathbf{0.79} \pm 0.001$ & $\underline{\emph{0.77}} \pm 0.001$ \\
         30\% & $0.39 \pm 0.003$ & $0.52 \pm 0.012$ & $0.51 \pm 0.010$ & $0.42 \pm 0.002$ & $0.42 \pm 0.008$ & $0.55 \pm 0.002$ & $0.46 \pm 0.011$ & $\mathbf{0.80} \pm 0.001$ & $\underline{\emph{0.78}} \pm 0.004$ \\
         40\% & $0.54 \pm 0.008$ & $0.55 \pm 0.008$ & $0.56 \pm 0.008$ & $0.48 \pm 0.003$ & $0.46 \pm 0.004$ & $0.60 \pm 0.002$ & $0.58 \pm 0.010$ & $0.73 \pm 0.001$ & $\mathbf{0.74} \pm 0.002$ \\
         45\% & $0.55 \pm 0.007$ & $0.55 \pm 0.008$ & $0.62 \pm 0.009$ & $0.48 \pm 0.003$ & $0.48 \pm 0.001$ & $0.56 \pm 0.004$ & $0.27 \pm 0.009$ & $0.63 \pm 0.001$ & $\mathbf{0.67} \pm 0.004$ \\
         \bottomrule 
    \end{tabular}
    }
\end{table}

\begin{table}[tbp]
    \centering
    \caption{F1-score under different feature-noise rates.}
    \label{tab:feature_noise_proportion}
    \resizebox{\textwidth}{!}{%
    \begin{tabular}{c|ccccccccc} 
        \toprule 
        \multirow{2}{*}{\shortstack{Noise\\Rate}} 
         & \multirow{2}{*}{LOO} & Data & Beta & Data & Influence & KNN & \multirow{2}{*}{AME} & Data & \multirow{2}{*}{NDDV} \\ 
         & & Shapley & Shapley & Banzhaf & Function & Shapley & & -OOB & \\
         \midrule 
         5\% & $0.09 \pm 0.007$ & $0.10 \pm 0.009$ & $0.10 \pm 0.007$ & $0.07 \pm 0.004$ & $0.10 \pm 0.003$ & $0.17 \pm 0.003$ & $0.09 \pm 0.012$ & $0.15 \pm 0.002$ & $\mathbf{0.30} \pm 0.006$ \\
         10\% & $0.18 \pm 0.007$ & $0.18 \pm 0.010$ & $0.18 \pm 0.009$ & $0.15 \pm 0.005$ & $0.15 \pm 0.003$ & $0.15 \pm 0.003$ & $0.18 \pm 0.010$ & $0.21 \pm 0.002$ & $\mathbf{0.28} \pm 0.003$ \\
         20\% & $0.33 \pm 0.005$ & $0.01 \pm 0.008$ & $0.01 \pm 0.008$ & $0.28 \pm 0.002$ & $0.30 \pm 0.002$ & $0.27 \pm 0.002$ & $0.01 \pm 0.010$ & $0.32 \pm 0.001$ & $\mathbf{0.34} \pm 0.003$ \\
         30\% & $0.43 \pm 0.008$ & $0.01 \pm 0.012$ & $0.01 \pm 0.010$ & $0.33 \pm 0.002$ & $0.35 \pm 0.008$ & $0.35 \pm 0.002$ & $0.01 \pm 0.012$ & $0.37 \pm 0.001$ & $\mathbf{0.45} \pm 0.005$ \\
         40\% & $0.51 \pm 0.008$ & $0.01 \pm 0.010$ & $0.01 \pm 0.008$ & $0.01 \pm 0.003$ & $0.37 \pm 0.004$ & $0.40 \pm 0.002$ & $0.01 \pm 0.010$ & $0.43 \pm 0.001$ & $\mathbf{0.57} \pm 0.004$ \\
         45\% & $0.53 \pm 0.007$ & $0.01 \pm 0.011$ & $0.01 \pm 0.009$ & $0.50 \pm 0.003$ & $0.47 \pm 0.001$ & $0.39 \pm 0.002$ & $0.01 \pm 0.012$ & $0.46 \pm 0.001$ & $\mathbf{0.62} \pm 0.006$ \\
         \bottomrule 
    \end{tabular}
    }
\end{table}

\subsection{Diagnostics and Sensitivity}

\noindent\textbf{Class-conditional detection diagnostics.}
These diagnostics use task labels as groups rather than protected attributes.
DTPRGap and DEOGap measure how evenly the fixed-budget corruption detector operates across classes.
The score threshold and corruption budget match Section~\ref{subsec:f1_score_comparison}.

\noindent\textbf{Local Threshold-Sensitivity Diagnostic.}
A group-mean score bound does not determine the behavior of a thresholded detector.
We estimate local TPR/FPR sensitivity to group-level score shifts and compare the resulting envelopes with measured gaps.
Table~\ref{tab:threshold_sensitivity_diagnostics} gives the diagnostic summary.
The envelopes are empirical local checks rather than direct consequences of Corollary~\ref{cor:score_gap_transfer}.

\begin{table}[tbp]
\centering
\caption{Class-wise detection diagnostics and gap bounds.}
\label{tab:threshold_sensitivity_diagnostics}
\resizebox{\textwidth}{!}{
\begin{tabular}{ccccccc}
\toprule
Dataset & $\varepsilon_U$ & $\hat L_{\mathrm{TPR}}$ & $\hat L_{\mathrm{FPR}}$ & \shortstack{Predicted\\DTPRGap bound} & \shortstack{Predicted\\DEOGap bound} & \shortstack{Measured NDDV\\DTPRGap / DEOGap} \\
\midrule
2dplanes    & $0.056$ & $1.8$ & $1.2$ & $0.101$ & $0.168$ & $0.041$ / $0.072$ \\
electricity & $0.048$ & $1.6$ & $1.1$ & $0.077$ & $0.130$ & $0.035$ / $0.058$ \\
BBC         & $0.072$ & $2.1$ & $1.4$ & $0.151$ & $0.252$ & $0.083$ / $0.139$ \\
IMDB        & $0.065$ & $1.9$ & $1.3$ & $0.124$ & $0.208$ & $0.062$ / $0.105$ \\
STL10       & $0.088$ & $2.3$ & $1.5$ & $0.202$ & $0.334$ & $0.117$ / $0.189$ \\
CIFAR10     & $0.094$ & $2.4$ & $1.6$ & $0.226$ & $0.376$ & $0.128$ / $0.214$ \\
\bottomrule
\end{tabular}
}
\end{table}

Across the six datasets, the measured gaps lie below the local linearized envelopes, typically at about $40\%$--$60\%$ of the bound.
The comparison is descriptive and becomes looser on the image data, which also show larger sampled comparison errors in Table~\ref{tab:epsilon_delta}.

\begin{figure*}[t]
    \centering
    \includegraphics[width=\textwidth,height=\textheight,keepaspectratio]{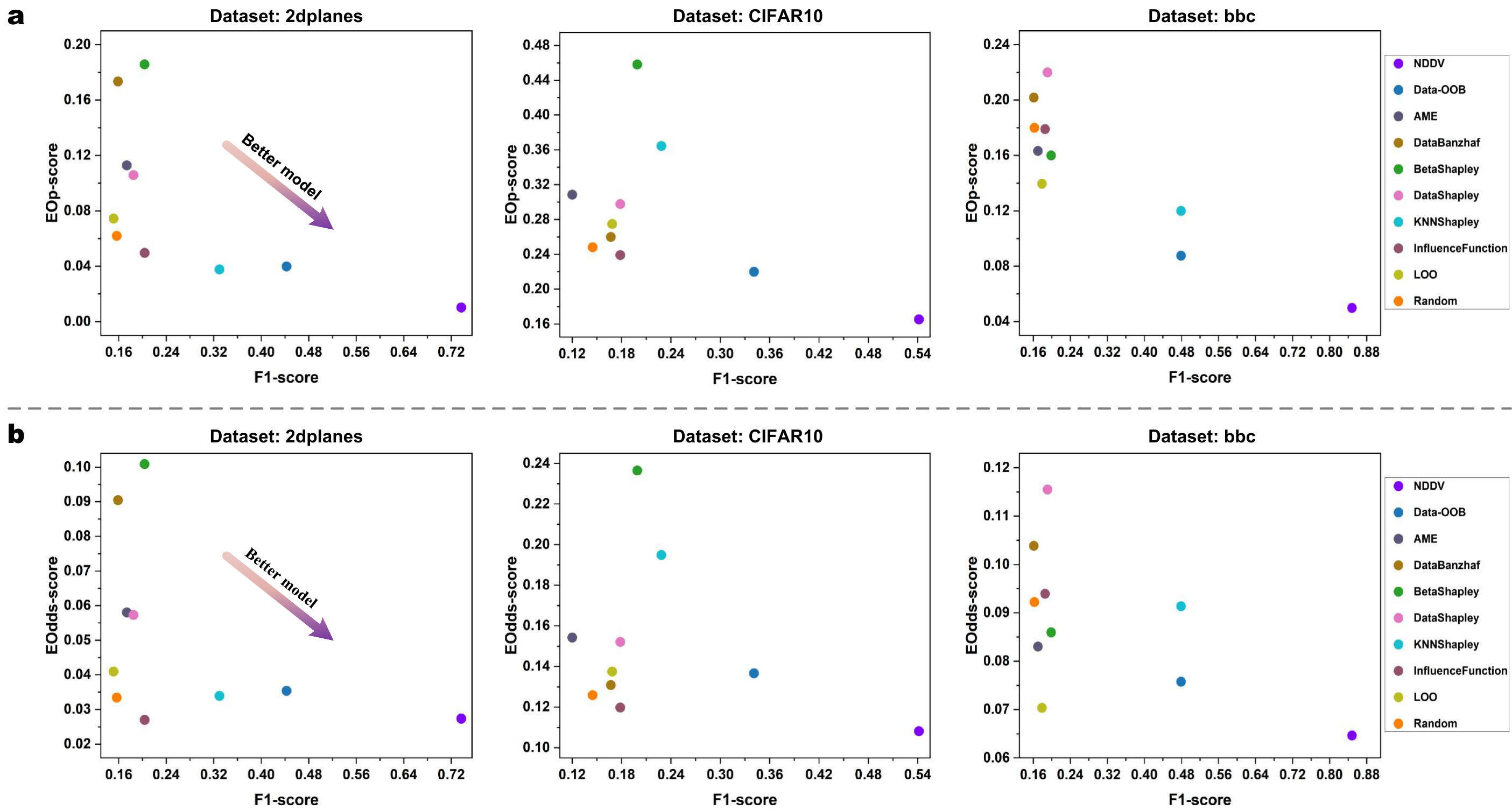}
    \vspace{-1em}
    \caption{Class-conditional detection diagnostics. a.
    DTPRGap evaluation. b.
    DEOGap evaluation.}
    \label{fig:class_conditional_diagnostics}
\end{figure*}

\begin{figure*}[t]
    \centering
    \includegraphics[width=\textwidth,height=0.72\textheight,keepaspectratio]{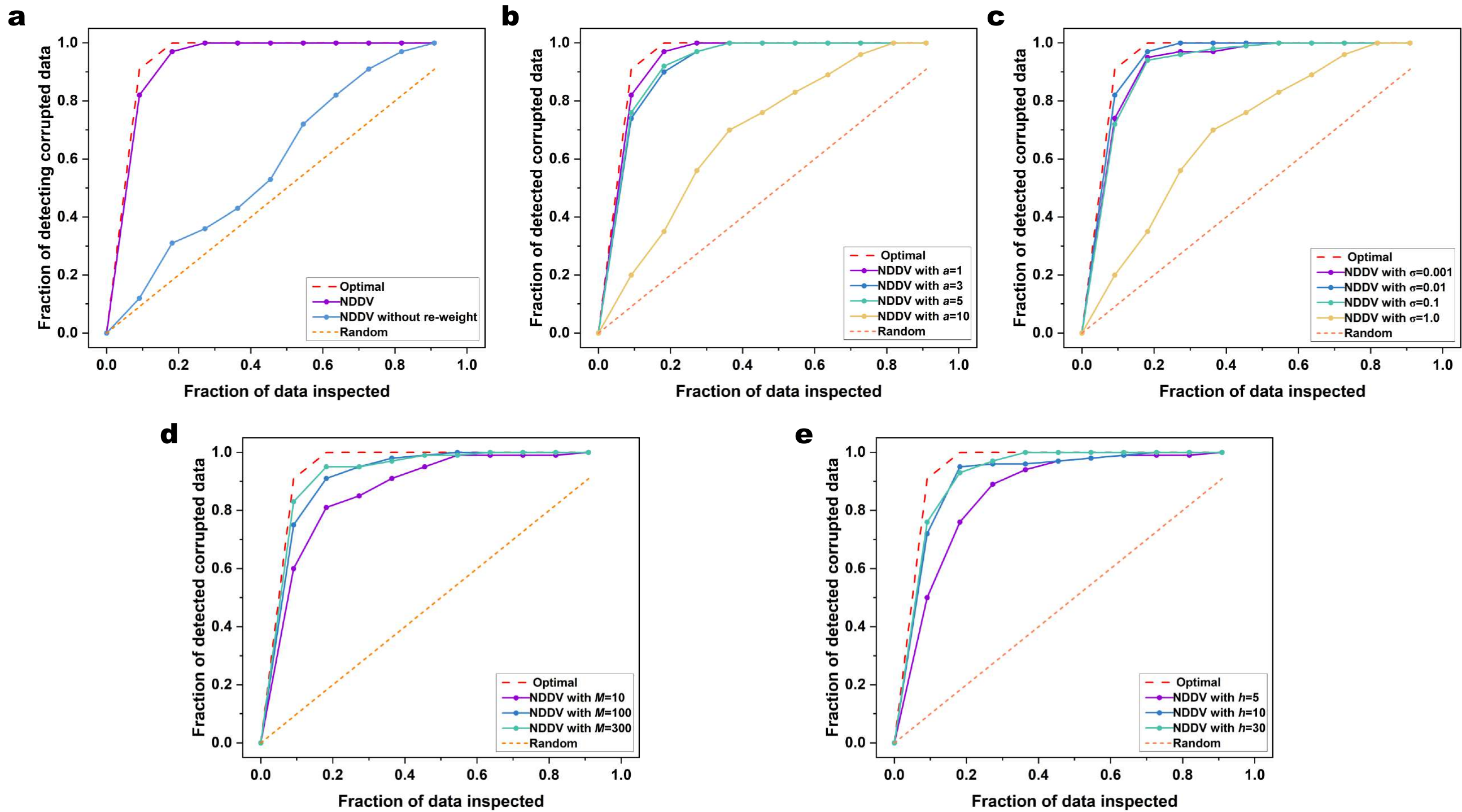}
    \vspace{-1em}
    \caption{NDDV sensitivity analysis on corrupted-sample detection. a.
    Sample reweighting. b.
    Mean-field interaction. c.
    Diffusion scale. d.
    Meta-set size. e.
    Meta-network width.}
    \label{fig:ablation_study}
\end{figure*}

Figure~\ref{fig:class_conditional_diagnostics} displays detection quality and class-conditional gaps on the same axes.
Panel~(a) of Fig.~\ref{fig:ablation_study} compares learned reweighting with the unweighted aggregate on 2dplanes, while Appendix~\ref{app:protected_attribute_validation} gives the corresponding weighted--unweighted comparison on Adult.
These checks show that the weights can alter the fitted score and its detection behavior, but they do not establish a monotone reduction of group gaps across datasets.

\begin{figure*}[t]
    \centering
    \includegraphics[width=\textwidth,height=0.72\textheight,keepaspectratio]{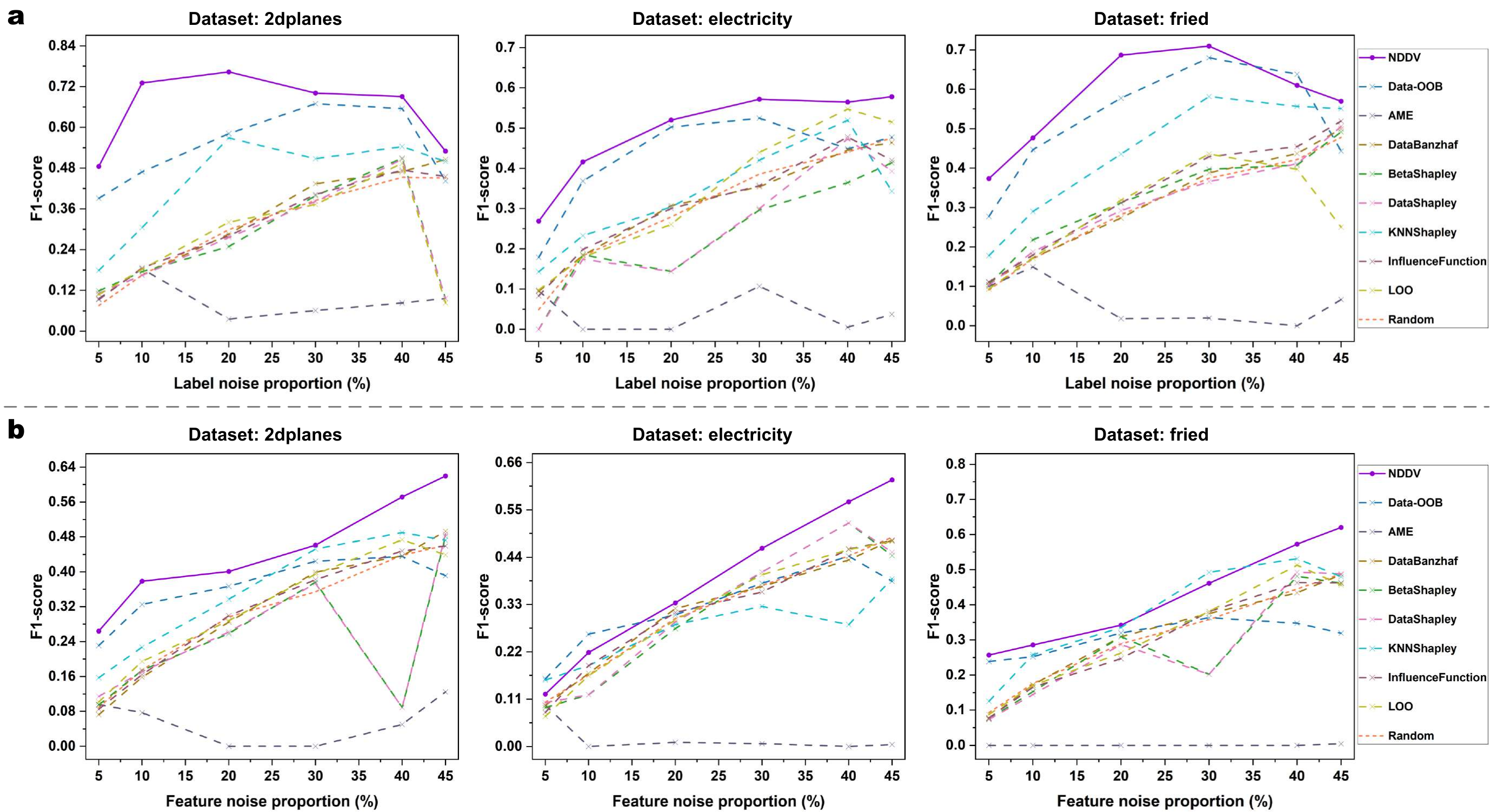}
    \vspace{-1em}
    \caption{Noisy data detection for the structured NDDV variant. a.
    Noisy-label detection. b.
    Noisy-feature detection.
    F1-score across six noise levels.}
    \label{fig:noise_interpretability_exp}
\end{figure*}

\vspace{0.5em}
\noindent\textbf{Noise-level diagnostics.}
Figure~\ref{fig:noise_interpretability_exp} summarizes the structured NDDV realization under increasing label and feature corruption.
The curves show stable performance across the tested range, with the largest relative gains in several feature-noise settings.
They also show that method rankings vary with the noise regime.

\vspace{0.5em}
\noindent\textbf{Sensitivity analysis.}
\label{app:sensitivity_analysis}

Figure~\ref{fig:ablation_study} varies one design choice at a time on the mislabeled-sample task: sample reweighting, mean-field strength, diffusion scale, meta-set size, and meta-network width.
The goal is to identify stable operating ranges and visible failure modes, not to decompose the causal effect of each architectural component.

The main trends are stable across panels.
Removing the learned weights lowers the detection and manipulation curves.
Moderate mean-field interaction and small diffusion are robust, whereas $a=10$ or $\sigma=1.0$ visibly degrades performance.
The meta-set size has little effect in this benchmark, while a very small hidden width is consistent with underfitting of the weight map.

Overall, the sensitivity curves support the operating range used in the main experiments and expose the expected failure modes: excessive coupling, excessive diffusion, and insufficient weight-network capacity.

\subsection{Recent Scalable Baselines}
\label{subsec:recent_valuation_methods_comparison}

To connect the main benchmarks with recent scalable valuation estimators, we include a complementary comparison against recent OpenDataVal-style estimators, GhostSuite, and LossVal.
Table~\ref{tab:recent_valuation_methods_comparison} uses a small-tabular protocol with different datasets, sample sizes, and noise settings.
The valuation-fairness rows use the composite score from the small-tabular semivalue-calibration protocol.
NDDV improves over the strongest recent baseline in each evaluated valuation-fairness, feature-noise, and label-noise setting.

\begin{table*}[t]
    \centering
    \caption{Comparison with recent valuation methods.}
    \label{tab:recent_valuation_methods_comparison}
    \resizebox{0.8\textwidth}{!}{%
    \begin{tabular}{cccccc}
        \toprule
        Dataset & Task & Best baselines & Baseline & NDDV & $\Delta$ \\
        \midrule
        2dplanes & Fairness & CS-Shapley & 10.357 & \textbf{16.341} & \textbf{+5.984} \\
        pol & Fairness & VolumeShapley & 9.509 & \textbf{16.320} & \textbf{+6.812} \\
        fire & Fairness & CS-Shapley & 8.820 & \textbf{16.318} & \textbf{+7.498} \\
        \midrule
        2dplanes & Feature noise & SingularOOB & 0.453 & \textbf{0.632} & \textbf{+0.179} \\
        pol & Feature noise & SingularLAVA & 0.656 & \textbf{0.821} & \textbf{+0.165} \\
        fire & Feature noise & SingularLAVA & 0.520 & \textbf{0.640} & \textbf{+0.119} \\
        \midrule
        2dplanes & Label noise & GhostSuite & 0.495 & \textbf{0.616} & \textbf{+0.122} \\
        pol & Label noise & SingularOOB & 0.485 & \textbf{0.595} & \textbf{+0.110} \\
        fire & Label noise & GhostSuite & 0.470 & \textbf{0.591} & \textbf{+0.121} \\
        \bottomrule
    \end{tabular}
    }
\end{table*}

Eigen-Value is retained only in the small-tabular comparison because its eigenvalue-based computation is not used in the larger ImageNet100 scaling study.

\begin{figure*}[t]
    \centering
    \includegraphics[width=\textwidth,keepaspectratio]{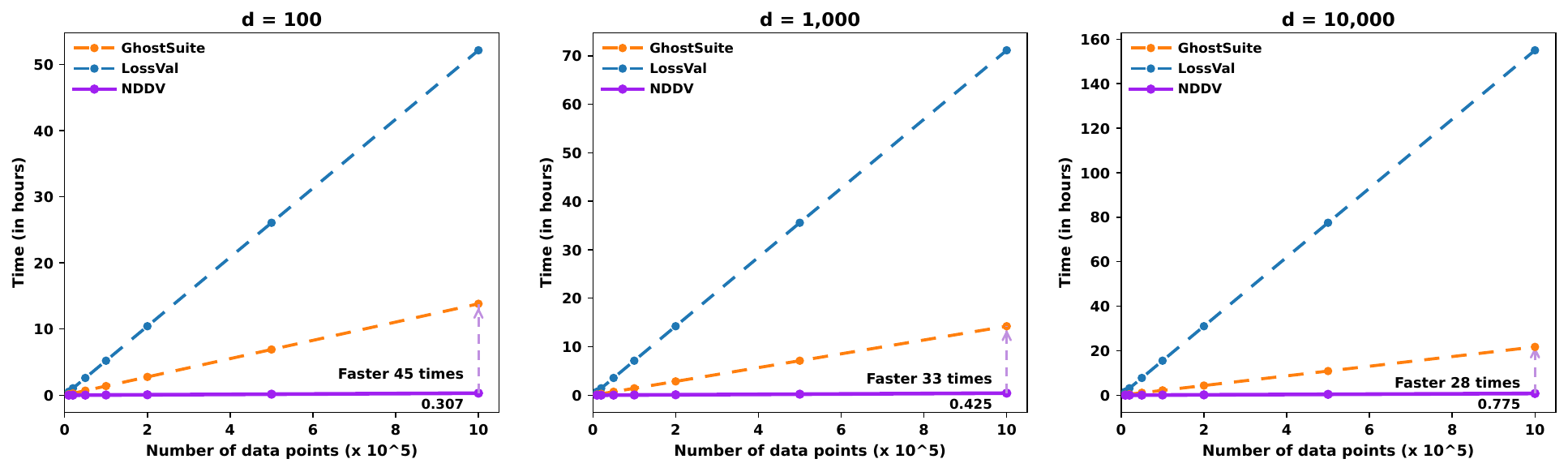}
    \vspace{-1em}
    \caption{Runtime scaling of NDDV, GhostSuite, and LossVal as the number of training data points grows, shown for feature dimensions $d=100$, $1{,}000$, and $10{,}000$.}
    \label{fig:runtime_comparison_new}
\end{figure*}

\vspace{0.5em}
\noindent\textbf{Large-scale runtime comparison.}
Figure~\ref{fig:runtime_comparison_new} compares NDDV with GhostSuite and LossVal as the sample size grows toward $10^6$.
Under the same synthetic scaling protocol, NDDV remains the fastest method across the three feature dimensions, with a $28$--$45\times$ extrapolated speedup over LossVal at the largest scale.

\begin{figure*}[t]
    \centering
    \includegraphics[width=.85\textwidth,keepaspectratio]{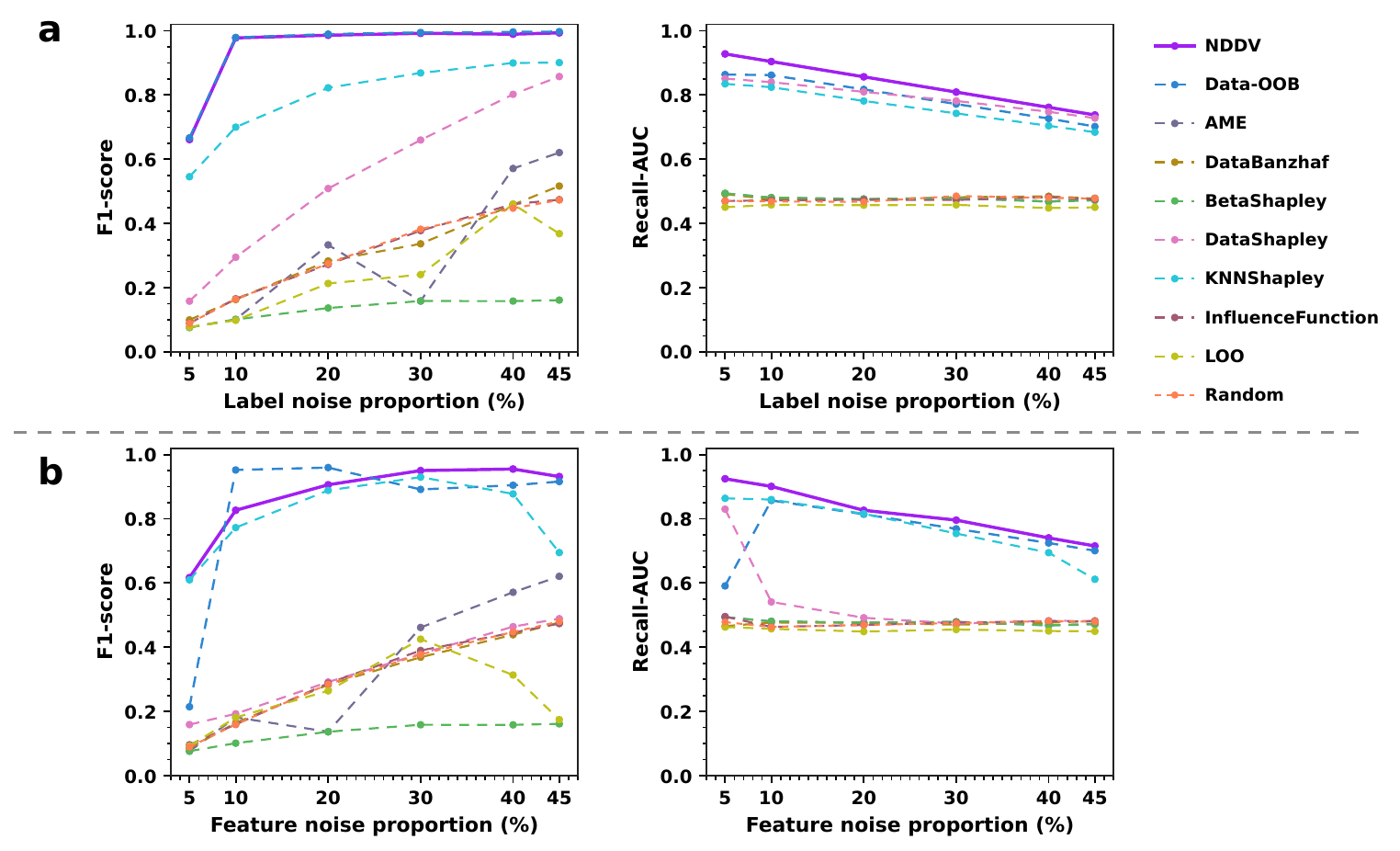}
    \vspace{-1em}
    \caption{ImageNet100 F1-score and recall-AUC curves under label and feature noise.}
    \label{fig:imagenet100_f1_recall}
\end{figure*}

Figure~\ref{fig:imagenet100_f1_recall} shows the corresponding ImageNet100 corrupted-sample detection curves under label and feature noise.
NDDV achieves higher F1-score curves in the feature-noise regime and comparable recall-AUC curves across the evaluated settings.

Together, the recent-estimator and ImageNet100 results complement the classical semivalue comparisons by testing NDDV against newer scalable baselines under different sample-size regimes.

\vspace{0.5em}
\noindent\textbf{Scope and limitations.}
The empirical study fixes the representation and downstream predictor so that the valuation rule is the object being compared.
This design makes retraining-based checks feasible, but it does not guarantee the same numerical ordering under end-to-end representation learning or substantially larger predictive models.
Runtime results are tied to the stated architectures, budgets, and hardware.
The finite-marginal audit is conservative and percentile based.
A full validation of Corollary~\ref{cor:sampled_pairwise_certificate} would require pair-level certificates and multi-trajectory convergence checks.

The class-conditional metrics are diagnostic.
They compare corruption-detection behavior across task labels, not demographic fairness, and the Adult protected-attribute check in Appendix~\ref{app:protected_attribute_validation} is deliberately limited.
The standard deviations quantify variation under the stated protocol rather than robustness across model classes or data-generation mechanisms.

\section{Conclusion}
\label{sec:conclusion}

This paper introduced NDDV, a one-run dynamic data valuation framework that estimates the importance of data points from a stochastic state--adjoint trajectory rather than from repeated coalition retraining.
The method records coupled forward dynamics, propagates an exact pathwise adjoint for the sampled frozen-aggregate Euler system, and applies a mass-preserving calibration to produce relative marginal-contribution scores.
The analysis separates this trajectory-conditioned estimator from classical coalition values by quantifying stability and discretization errors and by relating local sample-weight sensitivities to finite add-one marginals through an explicit error bridge.
Across the fixed experimental protocols, NDDV releases scores at low measured cost and achieves competitive results on marginal-fidelity checks, corrupted-sample detection, noise robustness, and recent scalable-baseline comparisons.
These findings support dynamic state--adjoint trajectories as a practical alternative when the target is contribution within a realized training run.
The comparison with LOO, Shapley, and other semi-values remains conditional on the local-to-finite approximation regime, and the class-conditional and protected-attribute analyses should be interpreted as diagnostics.
Future work should strengthen pair-level certificates, incorporate path-integrated sensitivities, and extend the framework beyond fixed representations and predictors.

\acks{This work was supported by the National Natural Science Foundation of China (12202157), the Exploration Foundation of the Key Laboratory of CNC Equipment Reliability, Ministry of Education, and the National Key Laboratory of Automotive Chassis Integration and Bionics at Jilin University.
The authors declare no competing interests.}

\appendix

\section{Stationarity of the First-Order Reweighting Surrogate}
\label{app:first_order_meta_reweighting_convergence}
The meta-weighted aggregate uses the first-order objective in Eq.~\eqref{eqn:first_order_penalty}.
Since the inner solution, aggregate path, and discrete adjoint are approximate, the update is analyzed as a biased stochastic gradient.

Fix one outer stage, write $z=(\theta,\psi)$, and define
\begin{equation}\label{eqn:first_order_surrogate_appendix}
\mathcal J(z)
=
\frac1{M_{\mathrm{meta}}}\sum_{i=1}^{M_{\mathrm{meta}}}\ell_i(\psi)
+\lambda\left[g(\theta,\psi)-g(\theta,\hat\psi)\right].
\end{equation}
The stochastic update is
\begin{equation}\label{eqn:first_order_update_appendix}
z^{k+1}=z^k-\eta_kG_k.
\end{equation}

\begin{assumption}[Biased stochastic-gradient regularity]
\label{assump:first_order_surrogate}
The objective is bounded below by $\mathcal J_{\inf}$ and has an $L_{\mathcal J}$-Lipschitz gradient.
Conditional on $z^k$,
\[
\mathbb E[G_k\mid z^k]=\nabla\mathcal J(z^k)+b_k,
\qquad
\|b_k\|\le\beta_k,
\]
and
\[
\mathbb E\!\left[
\|G_k-\mathbb E[G_k\mid z^k]\|^2
\,\middle|\,z^k
\right]
\le\sigma_G^2.
\]
\end{assumption}

\begin{proposition}[Stationarity with an Approximation-Bias Floor]
\label{prop:first_order_meta_reweighting_stationarity}
Under Assumption~\ref{assump:first_order_surrogate}, let $0<\eta_k\le1/(4L_{\mathcal J})$.
Then
\begin{equation}\label{eqn:biased_stationarity_bound}
\frac{\sum_{k=0}^{K-1}\eta_k\,
\mathbb E\|\nabla\mathcal J(z^k)\|^2}
{\sum_{k=0}^{K-1}\eta_k}
\le
\frac{2(\mathcal J(z^0)-\mathcal J_{\inf})}
{\sum_{k=0}^{K-1}\eta_k}
+4\frac{\sum_{k=0}^{K-1}\eta_k\beta_k^2}
{\sum_{k=0}^{K-1}\eta_k}
+L_{\mathcal J}\sigma_G^2
\frac{\sum_{k=0}^{K-1}\eta_k^2}
{\sum_{k=0}^{K-1}\eta_k}.
\end{equation}
For the constant choice $\eta_k=\eta=\min\{1/(4L_{\mathcal J}),c/\sqrt K\}$,
\begin{equation}\label{eqn:biased_stationarity_rate}
\min_{0\le k<K}
\mathbb E\|\nabla\mathcal J(z^k)\|^2
\le
O(K^{-1/2})
+4\overline\beta_K^2,
\qquad
\overline\beta_K^2=\frac1K\sum_{k=0}^{K-1}\beta_k^2.
\end{equation}
The usual nonconvex $O(K^{-1/2})$ term is recovered when the approximation bias vanishes.
With nonzero bias, the limiting neighborhood is controlled by the inner-solution, mean-field, and discretization errors.
\end{proposition}

\begin{corollary}[Bias decomposition for the NDDV update]
\label{cor:nddv_gradient_bias}
Suppose the surrogate gradient is Lipschitz with respect to the inner solution, the aggregate path, and the adjoint trajectory.
Let $r_k=\|\hat\psi^k-\psi^*(\theta^k)\|$, let $\eta_{\mu,k}+\Gamma_{\mathrm{law},k}$ be the model-to-frozen discrepancy from Proposition~\ref{prop:frozen_mean_field_stability}, and let $\epsilon_{\Delta,k}$ be the discrete-adjoint error.
Then constants $C_{\mathrm{in}},C_{\mathrm{MF}},C_\Delta$ exist such that
\begin{equation}\label{eqn:nddv_gradient_bias_decomposition}
\beta_k
\le
C_{\mathrm{in}}r_k
+C_{\mathrm{MF}}(\eta_{\mu,k}+\Gamma_{\mathrm{law},k})
+C_\Delta\epsilon_{\Delta,k}.
\end{equation}
Under the Euler assumptions of Corollary~\ref{cor:numerical_score_error}, $\epsilon_{\Delta,k}=O(\Delta t^{1/2})$.
Substitution into Eq.~\eqref{eqn:biased_stationarity_bound} expresses the optimization floor in the approximation terms used in the valuation analysis.
\end{corollary}

\begin{proof}[Proof of Proposition~\ref{prop:first_order_meta_reweighting_stationarity}]
Write $g_k=\nabla\mathcal J(z^k)$ and $G_k=g_k+b_k+\xi_k$, where $\mathbb E[\xi_k\mid z^k]=0$ and $\mathbb E[\|\xi_k\|^2\mid z^k]\le\sigma_G^2$.
Smoothness gives
\begin{align*}
\mathbb E[\mathcal J(z^{k+1})\mid z^k]
&\le
\mathcal J(z^k)
-\eta_k\|g_k\|^2
-\eta_k\langle g_k,b_k\rangle\\
&\quad+
\frac{L_{\mathcal J}\eta_k^2}{2}
\left(\|g_k+b_k\|^2+\sigma_G^2\right).
\end{align*}
Using $-\langle g_k,b_k\rangle\le\|g_k\|^2/4+\|b_k\|^2$, $\|g_k+b_k\|^2\le2\|g_k\|^2+2\|b_k\|^2$, and $\eta_k\le1/(4L_{\mathcal J})$ yields
\[
\frac{\eta_k}{2}\|g_k\|^2
\le
\mathcal J(z^k)-
\mathbb E[\mathcal J(z^{k+1})\mid z^k]
+2\eta_k\beta_k^2 +\frac{L_{\mathcal J}\eta_k^2}{2}\sigma_G^2.
\]
Taking expectations, summing over $k$, and using $\mathcal J(z^K)\ge\mathcal J_{\inf}$ gives Eq.~\eqref{eqn:biased_stationarity_bound}.
The constant-stepsize statement follows by substitution.
\end{proof}

\begin{proof}[Proof of Corollary~\ref{cor:nddv_gradient_bias}]
Add and subtract the surrogate gradient evaluated at the exact inner solution, population aggregate, and continuous adjoint.
Lipschitz continuity and the triangle inequality give Eq.~\eqref{eqn:nddv_gradient_bias_decomposition}.
The Euler rate follows from the strong discretization bound used in Corollary~\ref{cor:numerical_score_error}.
\end{proof}

\section{Pathwise Adjoint and SMP Projection}
\label{app:data_state_utility}

\begin{proof}[Proof of Theorem~\ref{thm:general_adjoint_perturbation}]
Let $\dot X_{i,t}=\left.\partial_\epsilon X_{i,t}^{\epsilon}\right|_{\epsilon=0}$.
Differentiating Eq.~\eqref{eqn:general_perturbed_state} gives the variational equation
\begin{equation}\label{eqn:general_variational_state}
\mathrm d\dot X_{i,t}
=\left[
\nabla_x b(X_{i,t},\mu_t,\psi_t)\dot X_{i,t}
+h_{i,t}(X_{i,t})
\right]\mathrm dt,
\qquad
\dot X_{i,0}=\xi_i.
\end{equation}
Differentiating the objective under the expectation yields
\begin{align}
\mathcal J_i'(0)
&=\mathbb E\!\left[
\int_0^T\!\left(
\nabla_xR_i(X_{i,t},\mu_t,\psi_t)\!\cdot\!\dot X_{i,t}
+r_{i,t}(X_{i,t})
\right)\mathrm dt \right.\nonumber\\
&\qquad\left.
+\nabla_x\Phi_i(X_{i,T},\mu_T,\psi_T)\!\cdot\!\dot X_{i,T}
+q_i(X_{i,T})
\right].
\label{eqn:direct_first_variation}
\end{align}
Apply It\^o's product rule to $Y_{i,t}\cdot\dot X_{i,t}$.
Since $\dot X_i$ has finite variation, its quadratic covariation with the martingale part of $Y_i$ is zero.
Eqs.~\eqref{eqn:adjoint_eq} and~\eqref{eqn:general_variational_state} cancel the terms containing $\nabla_x b$, giving
\begin{equation}\label{eqn:adjoint_cancellation_identity}
\mathbb E\!\left[Y_{i,T}\cdot\dot X_{i,T}-Y_{i,0}\cdot\xi_i\right]
=
\mathbb E\int_0^T\!\left(
Y_{i,t}\cdot h_{i,t}(X_{i,t})
+\nabla_xR_i(X_{i,t},\mu_t,\psi_t)\cdot\dot X_{i,t}
\right)\mathrm dt.
\end{equation}
The terminal condition gives $\nabla_x\Phi_i(X_{i,T},\mu_T,\psi_T)=-Y_{i,T}$.
Substitution into Eq.~\eqref{eqn:direct_first_variation} proves Eq.~\eqref{eqn:general_adjoint_representation}.
For terminal rescaling, differentiating $\Phi_i((1+\epsilon)x,\mu_T,\psi_T)$ at $\epsilon=0$ gives $q_i(x)=x\cdot\nabla_x\Phi_i(x,\mu_T,\psi_T)$.
The remaining marginal-contribution terms vanish and yield Eq.~\eqref{eqn:utility}.
\end{proof}

\begin{proof}[Proof of Proposition~\ref{prop:frozen_mean_field_stability}]
Let $\delta X_t=X_{i,t}^\star-\bar X_{i,t}$.
Lipschitz continuity of the drift gives
\[
\|b(X_{i,t}^\star,\mu_t^\star,\psi_t)-b(\bar X_{i,t},\bar\mu_t,\psi_t)\|
\le L_x\|\delta X_t\|+L_\mu\|\mu_t^\star-\bar\mu_t\|.
\]
The diffusion terms cancel because the two systems are coupled with the same Wiener path.
SDE stability and Gronwall's lemma yield
\begin{equation}\label{eqn:app_state_mf_stability}
\|\delta X\|_{\mathcal S^2}\le C_X\eta_\mu.
\end{equation}

Set $\delta Y=Y_i^\star-\bar Y_i$ and $\delta Z=Z_i^\star-\bar Z_i$.
The terminal condition is Lipschitz in $(x,\mu)$, and the population adjoint contains the terminal law derivative.
Hence
\[
\|\delta Y_T\|_{L^2}
\le C_T(\|\delta X_T\|_{L^2}+\eta_\mu)
+\|\mathfrak M_{i,T}^{\Phi}\|_{L^2}.
\]
Subtract the two BSDEs.
Their driver difference is bounded by
\[
L_X\|\delta X_t\|+L_\mu\|\mu_t^\star-\bar\mu_t\| +L_Y\|\delta Y_t\|+L_Z\|\delta Z_t\| +\|\mathfrak M_{i,t}\|.
\]
Applying It\^o's formula to $e^{\beta t}\|\delta Y_t\|^2$, choosing $\beta$ to absorb the $Y$ and $Z$ terms, and using the Burkholder--Davis--Gundy inequality gives
\begin{equation}\label{eqn:app_bsde_mf_stability}
\|\delta Y\|_{\mathcal S^2}+\|\delta Z\|_{\mathcal H^2}
\le C_Y\bigl(\|\delta X\|_{\mathcal S^2}+\eta_\mu+\Gamma_{\mathrm{law}}\bigr).
\end{equation}
Combining Eqs.~\eqref{eqn:app_state_mf_stability} and~\eqref{eqn:app_bsde_mf_stability} proves Eq.~\eqref{eqn:frozen_mean_field_stability}.

For the terminal score, write
\begin{align*}
|X_{i,T}^\star\cdot Y_{i,T}^\star-\bar X_{i,T}\cdot\bar Y_{i,T}|
&\le \|X_{i,T}^\star-\bar X_{i,T}\|\,\|Y_{i,T}^\star\|
+\|\bar X_{i,T}\|\,\|Y_{i,T}^\star-\bar Y_{i,T}\|.
\end{align*}
Cauchy--Schwarz and the uniform second-moment bounds give Eq.~\eqref{eqn:frozen_score_stability}.
The rate follows by substituting the aggregate and interaction bounds.
\end{proof}

\begin{proof}[Proof of Theorem~\ref{thm:discrete_adjoint_identity}]
For fixed $\bar\mu_s$, $\psi_s$, and $\Delta W_{i,s}$, the continuation objective satisfies
\begin{equation}\label{eqn:discrete_bellman_identity}
J_{i,s}^{\Delta}(x)
=R_i(x,\bar\mu_s,\psi_s)\Delta t
+J_{i,s+1}^{\Delta}(F_{i,s}(x)).
\end{equation}
At $s=S$, $J_{i,S}^{\Delta}(x)=\Phi_i(x,\bar\mu_S,\psi_S)$, hence $Y_{i,S}^{\Delta}=-\nabla_xJ_{i,S}^{\Delta}(X_{i,S}^{\Delta})$.
Assume the identity holds at step $s+1$.
Differentiating Eq.~\eqref{eqn:discrete_bellman_identity} yields
\begin{align*}
\nabla_xJ_{i,s}^{\Delta}(X_{i,s}^{\Delta})
&=\nabla_xR_i(X_{i,s}^{\Delta},\bar\mu_s,\psi_s)\Delta t\\
&\quad+
\left[I+\nabla_xb(X_{i,s}^{\Delta},\bar\mu_s,\psi_s)\Delta t\right]^\top
\nabla_xJ_{i,s+1}^{\Delta}(X_{i,s+1}^{\Delta}).
\end{align*}
Substituting the induction hypothesis and multiplying by $-1$ gives
\begin{align*}
-\nabla_xJ_{i,s}^{\Delta}(X_{i,s}^{\Delta})
&=Y_{i,s+1}^{\Delta}
+\left[\nabla_xb(X_{i,s}^{\Delta},\bar\mu_s,\psi_s)^\top Y_{i,s+1}^{\Delta}
-\nabla_xR_i(X_{i,s}^{\Delta},\bar\mu_s,\psi_s)\right]\Delta t\\
&=Y_{i,s+1}^{\Delta}
+\nabla_x\mathcal H_i(X_{i,s}^{\Delta},\bar\mu_s,Y_{i,s+1}^{\Delta},0,\psi_s)\Delta t,
\end{align*}
which is the implemented recursion.
Backward induction proves Eq.~\eqref{eqn:discrete_adjoint_identity}.
Eq.~\eqref{eqn:discrete_directional_identity} follows from the directional derivative definition.
\end{proof}

\begin{proof}[Justification of the conditional projection in Eq.~\eqref{eqn:conditional_projection}]
Under Assumption~\ref{assump:adjoint_regular}, Eq.~\eqref{eqn:pathwise_adjoint} has a unique square-integrable pathwise solution.
Because the diffusion is additive,
\[
\nabla_x\mathcal H_i(X_{i,s},\mu_s,y,0,\psi_s)
=\nabla_xb(X_{i,s},\mu_s,\psi_s)^\top y -\nabla_xR_i(X_{i,s},\mu_s,\psi_s),
\]
is affine in $y$.
Taking conditional expectation in Eq.~\eqref{eqn:pathwise_adjoint}, applying the tower property, and using $\mathcal F_s$-measurability gives
\begin{align*}
Y_{i,t}
&=\mathbb E\!\left[
-\nabla_x\Phi_i(X_{i,T},\mu_T,\psi_T)
+\int_t^T
\nabla_x\mathcal H_i(X_{i,s},\mu_s,Y_{i,s},0,\psi_s)\,\mathrm ds
\,\middle|\,\mathcal F_t\right].
\end{align*}
The martingale representation theorem provides a unique square-integrable process $Z_{i,t}$ such that
\[
Y_{i,t} =-\nabla_x\Phi_i(X_{i,T},\mu_T,\psi_T) +\int_t^T\nabla_x\mathcal H_i(X_{i,s},\mu_s,Y_{i,s},Z_{i,s},\psi_s)\,\mathrm ds -\int_t^TZ_{i,s}\,\mathrm dW_{i,s}.
\]
Since $\Sigma$ is independent of the state, the $Z$ term does not enter $\nabla_x\mathcal H_i$.
The last display is the integral form of Eq.~\eqref{eqn:adjoint_eq}.
\end{proof}

\begin{proof}[Proof of Corollary~\ref{cor:numerical_score_error}]
Let $\bar U_i^{M_{\mathrm{traj}},\Delta}=M_{\mathrm{traj}}^{-1}\sum_{m=1}^{M_{\mathrm{traj}}}h_i(X_{i,S}^{\Delta,(m)})$.
Add and subtract $\mathbb E[h_i(X_{i,S}^{\Delta})]$.
Minkowski's inequality gives
\begin{align*}
\|\bar U_i^{M_{\mathrm{traj}},\Delta}-\mathbb E[h_i(X_{i,T})]\|_{L^2}
&\le
\|\bar U_i^{M_{\mathrm{traj}},\Delta}-\mathbb E[h_i(X_{i,S}^{\Delta})]\|_{L^2}\\
&\quad+
|\mathbb E[h_i(X_{i,S}^{\Delta})-h_i(X_{i,T})]|.
\end{align*}
Independence of the $M_{\mathrm{traj}}$ paths bounds the first term by $\sigma_{h,i}/\sqrt{M_{\mathrm{traj}}}$.
Lipschitz continuity and the strong Euler bound control the second by $L_{h,i}C_{X,i}\Delta t^{1/2}$.
\end{proof}

\section{Sample-Weight Marginals and Semi-Values}
\label{app:ranking_consistency}

\begin{proof}[Proof of Proposition~\ref{prop:continuous_marginal_identity}]
Absolute continuity and the endpoint conditions give
\[
U(C\cup\{i\})-U(C) =U_{C,i}(1)-U_{C,i}(0) =\int_0^1m_i(C,\lambda)\,\mathrm d\lambda.
\]
If $m_i(C,\cdot)$ is $L_{i,C}$-Lipschitz, then
\begin{align*}
\left|\int_0^1m_i(C,\lambda)\,\mathrm d\lambda-m_i(C,\lambda_0)\right|
&\le
\int_0^1|m_i(C,\lambda)-m_i(C,\lambda_0)|\,\mathrm d\lambda\\
&\le
L_{i,C}\int_0^1|\lambda-\lambda_0|\,\mathrm d\lambda\\
&=
\frac{L_{i,C}}2\left(\lambda_0^2+(1-\lambda_0)^2\right).
\end{align*}
\end{proof}

\begin{proof}[Proof of Proposition~\ref{prop:local_adjoint_calibration}]
Differentiating Eq.~\eqref{eqn:utility_calibration_decomposition} at $\lambda_0$ gives
\[
m_i(C,\lambda_0) =\Gamma_{C,i}'(\lambda_0) +\nabla g_i(X_{i,T}^{C,\lambda_0})^\top
\partial_\lambda X_{i,T}^{C,\lambda_0}.
\]
Substitute Eq.~\eqref{eqn:inclusion_tangent_decomposition}:
\[
m_i(C,\lambda_0) =\Gamma_{C,i}'(\lambda_0) +\rho_i(C,\lambda_0)u_i^{C,\lambda_0} +\nabla g_i(X_{i,T}^{C,\lambda_0})^\top r_i(C,\lambda_0).
\]
Subtract $u_i^{C,\lambda_0}$ and apply the triangle and Cauchy--Schwarz inequalities.
Replacing $u_i^{C,\lambda_0}$ by $s_i(\mathcal T)$ adds $\varepsilon_i^{\mathrm{ctx}}$ by another triangle inequality.
\end{proof}

\begin{proof}[Proof of Theorem~\ref{thm:continuous_inclusion_bridge}]
Insert $m_i(C,\lambda_0)$ between the finite coalition marginal and $s_i(\mathcal T)$:
\begin{align*}
&\left|s_i(\mathcal T)-[U(C\cup\{i\})-U(C)]\right|\\
&\qquad\le
|s_i(\mathcal T)-m_i(C,\lambda_0)|
+
\left|m_i(C,\lambda_0)-\int_0^1m_i(C,\lambda)\,\mathrm d\lambda\right|.
\end{align*}
Proposition~\ref{prop:local_adjoint_calibration} bounds the first term, and Eq.~\eqref{eqn:local_to_finite_marginal} bounds the second.
\end{proof}

\subsection{Pairwise Decomposition of a Symmetric Semi-Value}
Throughout this subsection, sums over $A$ range over subsets of $[N]\setminus\{i,j\}$.
For $A\subseteq[N]\setminus\{i,j\}$, split the coalitions in $\phi_{\omega,i}$ into $A$ and $A\cup\{j\}$, and those in $\phi_{\omega,j}$ into $A$ and $A\cup\{i\}$.
Writing $k=|A|$ gives
\begin{align}
\phi_{\omega,i}(U)-\phi_{\omega,j}(U)
&=\sum_{A\subseteq[N]\setminus\{i,j\}}
\left(p_k^\omega+p_{k+1}^\omega\right)
\left[U(A\cup\{i\})-U(A\cup\{j\})\right]\nonumber\\
&=\sum_Ac_A^\omega\left[\Delta_i(A)-\Delta_j(A)\right].
\label{eqn:semivalue_pairwise_decomp}
\end{align}
The coefficients are nonnegative and sum to one:
\begin{align*}
\sum_Ac_A^\omega
&=\sum_{k=0}^{N-2}\binom{N-2}{k}
\left[
\frac{\omega_k}{\binom{N-1}{k}}
+
\frac{\omega_{k+1}}{\binom{N-1}{k+1}}
\right]\\
&=\sum_{k=0}^{N-2}\frac{N-1-k}{N-1}\omega_k
+\sum_{\ell=1}^{N-1}\frac{\ell}{N-1}\omega_\ell
=\sum_{k=0}^{N-1}\omega_k=1.
\end{align*}

\begin{proof}[Proof of Theorem~\ref{thm:ranking_consistency}]
By Eq.~\eqref{eqn:dynamic_dv},
\[
\frac{N-1}{N}(\phi_i^{\mathrm{NDDV}}-\phi_j^{\mathrm{NDDV}})
=s_i(\mathcal T)-s_j(\mathcal T).
\]
Combining this identity with Eq.~\eqref{eqn:semivalue_pairwise_decomp} gives
\begin{align*}
&\left|
s_i(\mathcal T)-s_j(\mathcal T)
-
(\phi_{\omega,i}(U)-\phi_{\omega,j}(U))
\right|\\
&\quad\le
\sum_Ac_A^\omega
\left|
[s_i(\mathcal T)-\Delta_i(A)]
-[s_j(\mathcal T)-\Delta_j(A)]
\right|\\
&\quad\le
\sum_Ac_A^\omega(e_i(A)+e_j(A))
=\mathcal E_{\omega,ij}.
\end{align*}
Equation~\eqref{eqn:pair_specific_theoretical_budget} follows from Theorem~\ref{thm:continuous_inclusion_bridge}.
If one compared gap has magnitude larger than $\mathcal E_{\omega,ij}$, the other cannot cross zero, so their signs agree.
\end{proof}

\begin{proof}[Proof of Corollary~\ref{cor:sampled_pairwise_certificate}]
Conditioning on the trained model, the variables $Z_m$ are independent, lie in $[0,B_{ij}^{\max}]$, and satisfy $\mathbb E[Z_m]=\mathcal E_{\omega,ij}$.
Hoeffding's inequality gives
\[
\mathbb P\!\left(
\mathcal E_{\omega,ij}
>
\widehat{\mathcal E}_{\omega,ij}+r
\right)
\le
\exp\!\left(-\frac{2M_{\mathrm{coal}}r^2}{(B_{ij}^{\max})^2}\right).
\]
Choosing $r=B_{ij}^{\max}\sqrt{\log(1/\alpha)/(2M_{\mathrm{coal}})}$ proves Eq.~\eqref{eqn:sampled_pairwise_certificate}.
The ordering statement follows from Theorem~\ref{thm:ranking_consistency}.
For $P$ fixed pairs, a union bound with per-pair failure level $\alpha/P$ gives the result.
\end{proof}

\subsection{LOO and Shapley Specializations}
\label{app:dynamic_dv}
For LOO, $\omega_{N-1}=1$ and the remaining weights vanish.
Eq.~\eqref{eqn:semivalue_pair_weights} assigns unit mass to $A=[N]\setminus\{i,j\}$, so
\begin{equation}\label{eqn:loo_pair_specific_bound}
\left|
\frac{N-1}{N}(\phi_i^{\mathrm{NDDV}}-\phi_j^{\mathrm{NDDV}})
-(\phi_{\mathrm{loo},i}-\phi_{\mathrm{loo},j})
\right|
\le e_i(A)+e_j(A).
\end{equation}
For Shapley, $\omega_k=1/N$, and Eq.~\eqref{eqn:ranking_bound} uses the corresponding distribution $c_A^{\mathrm{Shap}}$.
The resulting error weights each coalition according to its contribution to the Shapley pairwise difference.

\section{Relative-State Stability and Score Centering}
\label{app:relative_state_stability}

\begin{proof}[Proof of Theorem~\ref{thm:relative_state_contraction}]
For $D_{ij,t}=X_{i,t}-X_{j,t}$, subtracting Eq.~\eqref{eqn:lq_drift} for $i$ and $j$ gives
\[
\mathrm dD_{ij,t}=-aD_{ij,t}\,\mathrm dt+\Sigma\,\mathrm d(W_{i,t}-W_{j,t}).
\]
Hence
\[
D_{ij,t}=e^{-at}(x_i-x_j) +\int_0^te^{-a(t-s)}\Sigma\,\mathrm d(W_{i,s}-W_{j,s}).
\]
The stochastic integral has zero mean. It\^o's isometry gives
\[
\mathbb E\left\|\int_0^te^{-a(t-s)}\Sigma\,\mathrm d(W_{i,s}-W_{j,s})\right\|^2
=2\|\Sigma\|_F^2\int_0^te^{-2a(t-s)}\,\mathrm ds =\frac{1-e^{-2at}}{a}\|\Sigma\|_F^2,
\]
which proves Eq.~\eqref{eqn:pairwise_state_contraction}.

For $\bar D_{g,t}=\bar X_{g,t}-\bar X_t$, the same cancellation yields
\[
\mathrm d\bar D_{g,t}=-a\bar D_{g,t}\,\mathrm dt
+\Sigma\,\mathrm d(\bar W_{g,t}-\bar W_t).
\]
By independence, the quadratic variation of $\bar W_g-\bar W$ is $(n_g^{-1}-N^{-1})I\,\mathrm dt$.
It\^o's isometry gives Eq.~\eqref{eqn:group_mean_state_contraction}.
\end{proof}

\begin{proof}[Proof of Corollary~\ref{cor:score_gap_transfer}]
Use the double-average identity
\[
\bar u_g-\bar u
=\frac1{n_gN}\sum_{i\in G_g}\sum_{j=1}^N
\left[h_i(X_{i,T})-h_j(X_{j,T})\right].
\]
For each pair,
\begin{align*}
|h_i(X_{i,T})-h_j(X_{j,T})|
&\le
|h_i(X_{i,T})-h_i(X_{j,T})|
+|h_i(X_{j,T})-h_j(X_{j,T})|\\
&\le L_h\|X_{i,T}-X_{j,T}\|+\kappa_{ij}.
\end{align*}
Taking expectations, applying Cauchy--Schwarz, and using Eq.~\eqref{eqn:pairwise_state_contraction} gives Eq.~\eqref{eqn:dynamic_group_utility_bound}.
Equation~\eqref{eqn:score_disparity} follows from Lemma~\ref{lem:score_centering}.
\end{proof}

\begin{proof}[Proof of Lemma~\ref{lem:score_centering}]
Let $S=\sum_{j=1}^Nu_j=N\bar u$.
From Eq.~\eqref{eqn:dynamic_dv},
\[
\phi_i
=u_i-\frac{S-u_i}{N-1} =\frac{N}{N-1}(u_i-\bar u).
\]
Averaging over $i$ gives $\bar\phi=0$.
Averaging over $G_g$ proves Eq.~\eqref{eqn:group_score_identity}.
\end{proof}

\section{Adult Protected-Attribute Check}
\label{app:protected_attribute_validation}

The OpenDataVal benchmarks do not provide demographic protected attributes, so the main group analysis uses class labels.
Adult is used as a limited protected-attribute check for the score and corruption-detection quantities.
The experiment is descriptive and does not establish the dynamical bounds in Section~\ref{subsec:mean_field_stability}.


\subsection{Score-gap ablation}
For each run, we compute the raw trajectory-utility gap $\varepsilon_U=\max_g|\bar u_g-\bar u|$, the centered score gap $\hat D_\phi=\max_g|\bar\phi_g-\bar\phi|$, and the learned weight range $\Delta_{\mathcal V}=\max_i v_i-\min_i v_i$.
This check concerns valuation-score disparity, not downstream prediction disparity.
Table~\ref{tab:adult_thm2} compares the weighted model with the ablation $v_i\equiv1$.

\begin{table}[tbp]
\centering
\caption{Protected-attribute valuation-score gap on Adult.}
\label{tab:adult_thm2}
\begin{adjustbox}{max width=0.9\textwidth}
\begin{tabular}{ccccc}
\toprule
Model & $\varepsilon_U$ & $\tfrac{N}{N-1}\varepsilon_U$ & Measured $\hat D_\phi$ & $\Delta_{\mathcal V}$ \\
\midrule
NDDV (weighted mean-field) & $0.061$ & $0.061$ & $0.061$ & $0.34$ \\
NDDV (ablation: $v_i\equiv 1$)    & $0.098$ & $0.098$ & $0.098$ & $0$ \\
\bottomrule
\end{tabular}
\end{adjustbox}
\end{table}

The measured centered score gap matches the scaled raw gap, as implied by Lemma~\ref{lem:score_centering}.
The weighted model yields smaller raw trajectory-utility and centered valuation-score gaps than the unweighted ablation.
This comparison is not causal because the weights alter the common trajectory, terminal map, and fitted meta objective simultaneously.

\subsection{Corruption-Detection Gaps}
We evaluate Adult protected-attribute gaps under the main corrupted-sample detection budget.
NDDV applies group-wise score calibration followed by a prespecified group-balanced budget rule that uses protected-group sizes but no corruption labels.
Table~\ref{tab:adult_cor4} gives finite-difference sensitivity slopes, local gap envelopes, DTPRGap, and DEOGap.
The NDDV envelope uses Corollary~\ref{cor:score_gap_transfer}.
Baseline envelopes are empirical plug-in diagnostics.

\begin{table}[tbp]
\centering
\caption{Protected-attribute detection-gap evaluation on Adult.}
\label{tab:adult_cor4}
\resizebox{0.95\textwidth}{!}{%
\begin{tabular}{ccccc}
\toprule
Method & $\hat L_{\mathrm{TPR}}$ & $\hat L_{\mathrm{FPR}}$ & \shortstack{Predicted\\DTPRGap / DEOGap} & \shortstack{Measured\\DTPRGap / DEOGap} \\
\midrule
LOO                & $4.9$  & $3.2$ & $0.296$ / $0.494$ & $0.104$ / $0.138$ \\
BetaShapley        & $10.8$ & $2.9$ & $0.658$ / $0.838$ & $0.083$ / $0.085$ \\
DataBanzhaf        & $5.2$  & $3.6$ & $0.315$ / $0.533$ & $\mathbf{0.002}$ / $\mathbf{0.016}$ \\
AME                & $5.5$  & $3.6$ & $0.337$ / $0.554$ & $0.076$ / $0.094$ \\
Data-OOB           & $2.3$  & $0.4$ & $0.140$ / $0.162$ & $0.092$ / $0.110$ \\
DVRL               & $3.1$  & $2.3$ & $0.189$ / $0.330$ & $0.104$ / $0.133$ \\
InfluenceSubsample & $2.8$  & $4.1$ & $0.168$ / $0.417$ & $0.038$ / $0.087$ \\
KNNShapley         & $9.2$  & $3.2$ & $0.560$ / $0.753$ & $0.059$ / $0.065$ \\
LAVA               & $0.01$ & $0.06$ & $\mathbf{0.001}$ / $\underline{0.004}$ & $0.033$ / $0.064$ \\
DU-Shapley         & $0.3$  & $0.7$ & $\underline{0.019}$ / $0.064$ & $0.038$ / $0.044$ \\
GhostSuite         & $0.7$  & $0.4$ & $0.042$ / $0.065$ & $0.111$ / $0.141$ \\
LossVal            & $0.7$  & $0.2$ & $0.042$ / $0.053$ & $0.123$ / $0.145$ \\
DataShapley        & $11.4$ & $3.0$ & $0.694$ / $0.875$ & $0.052$ / $0.068$ \\
NDDV               & $0.01$ & $0.01$ & $\mathbf{0.001}$ / $\mathbf{0.001}$ & $\underline{0.009}$ / $\underline{0.028}$ \\
\bottomrule
\end{tabular}
}
\end{table}

NDDV has the second-lowest measured DTPRGap and DEOGap, behind DataBanzhaf, and the smallest local envelope at the table resolution.
These results are a limited detection-gap audit on one protected attribute and should not be read as a demographic-fairness or valuation-quality ranking.

\bibliography{sample}

\end{document}